\documentclass{article} 
\DeclareUnicodeCharacter{25B9}{\textbullet}
\usepackage[preprint]{colm2025_conference}

\usepackage{microtype}
\usepackage{float}

\usepackage{hyperref}
\usepackage{url}
\usepackage{booktabs}
\usepackage{graphicx}
\usepackage{amsmath}
\usepackage{xcolor}
\usepackage{arydshln}
\usepackage{caption}

\usepackage{algorithm}
\usepackage{algpseudocode}
\usepackage{lipsum}

\usepackage{lineno}

\usepackage{booktabs}
\usepackage{tikz}
\usepackage{pifont}  
\usepackage{wrapfig} 

\definecolor{darkblue}{rgb}{0, 0, 0.5}
\definecolor{DarkGreen}{RGB}{34, 139, 34}
\definecolor{DarkPurple}{RGB}{142, 124, 195}
\definecolor{DarkOrange}{RGB}{246, 178, 107}
\definecolor{LightGreen}{RGB}{80, 170, 80}

\newcommand{\Yes}{\textcolor{LightGreen}{\ding{51}}} 
\newcommand{\No}{\textcolor{red}{\ding{55}}}

\hypersetup{colorlinks=true, citecolor=darkblue, linkcolor=darkblue, urlcolor=darkblue}

\title{ElaLoRA: Elastic \& Learnable Low-Rank Adaptation for Efficient Model Fine-Tuning}

\author{Huandong Chang\footnotemark[1]~~\footnotemark[2],~~
Zicheng Ma\footnotemark[1],~~
Mingyuan Ma,~~
Zhenting Qi\footnotemark[2], \\
\textbf{Andrew Sabot,~~Hong Jiang,~~H.~T. Kung} \\
Harvard University, SEAS
}

\begin{document}

\ifcolmsubmission
\linenumbers
\fi

\maketitle

\renewcommand{\thefootnote}{\fnsymbol{footnote}}
\footnotetext[1]{Equal contribution. Code will be available at \url{https://github.com/HuandongChang/ElaLoRA}}       
\footnotetext[2]{Corresponds to \texttt{huandongchang@fas.harvard.edu}, \texttt{zhentingqi@fas.harvard.edu}}  
\renewcommand{\thefootnote}{\arabic{footnote}}

\begin{abstract}
Low-Rank Adaptation (LoRA) has become a widely adopted technique for fine-tuning large-scale pre-trained models with minimal parameter updates \citep{hu2022lora}. However, existing methods rely on fixed ranks or focus solely on either rank pruning or expansion, failing to adapt ranks dynamically to match the importance of different layers during training. In this work, we propose ElaLoRA, an adaptive low-rank adaptation framework that dynamically prunes and expands ranks based on gradient-derived importance scores. To the best of our knowledge, ElaLoRA is the first method that enables both rank pruning and expansion during fine-tuning. Experiments across multiple benchmarks demonstrate that ElaLoRA consistently outperforms existing PEFT methods across different parameter budgets. Furthermore, our studies validate that layers receiving higher rank allocations contribute more significantly to model performance, providing theoretical justification for our adaptive strategy. By introducing a principled and adaptive rank allocation mechanism, ElaLoRA offers a scalable and efficient fine-tuning solution, particularly suited for resource-constrained environments.

\end{abstract}

\section{Introduction}

Scaling laws of transformer-based Pre-trained Language Models (PLMs) \citep{Vaswani+2017,he2020deberta, liu2019roberta} suggest that increasing model size leads to improved generalization and task performance, which has driven the rapid expansion of model architectures \citep{kaplan2020scaling}, from 330M parameters in BERT \citep{devlin2019bert} to 1.5B in GPT-2 \citep{radford2019language}, 175B in GPT-3 \citep{brown2020language}, and 671B in DeepSeek \citep{bi2024deepseek}, highlighting the trend toward ever-larger pretrained models. Despite these advances, Large Language Models (LLMs) remain constrained by their knowledge boundaries, requiring fine-tuning to specialize in domain-specific applications and adapt to evolving datasets \citep{brown2020language, achiam2023gpt, gunter2024apple}. Traditionally, full fine-tuning has been the standard approach \citep{devlin2019bert, radford2019language}, which is nevertheless prohibitively expensive in terms of memory and computation.

To address the computational burden of full fine-tuning, Parameter-efficient fine-tuning (PEFT) methods have been developed  \citep{ding2023parameter}, with Low-Rank Adaptation (LoRA) \citep{hu2022lora} being a widely used approach that reduces trainable parameters without increasing inference latency. However, LoRA's fixed rank allocation leads to suboptimal performance by failing to account for layer-specific importance \citep{zhang2023adalora}. Dynamic rank allocation methods like AdaLoRA \citep{zhang2023adalora} and SaLoRA \citep{hu2023structure} decompose a matrix using singular value decomposition (SVD) and selectively prune its singular values to control the rank of the matrix, but these methods are computationally inefficient as they begin with a high rank. IncreLoRA \citep{zhang2023increlora} mitigates this by starting with a minimal rank and increasing it heuristically. However, early training samples may not be effectively learned or utilized when the rank is small.

\begin{figure}
    \centering
    \includegraphics[width=1.02\linewidth]{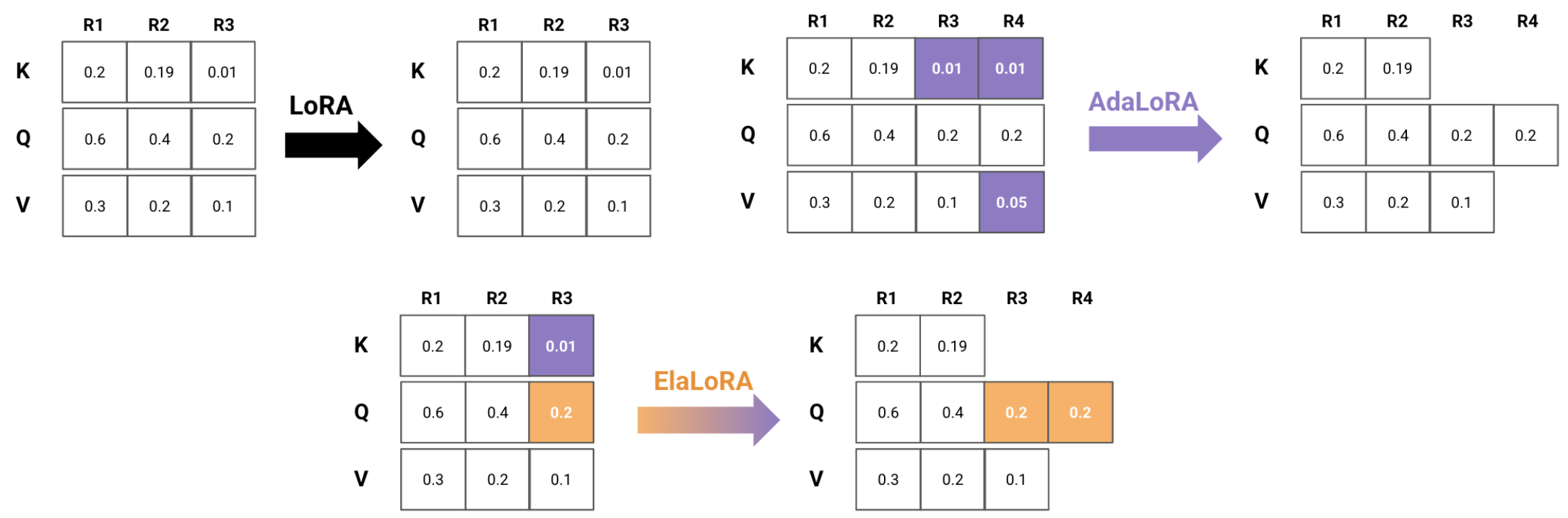}
        \caption{Comparison of ElaLoRA, LoRA, and AdaLoRA. LoRA has \textbf{fixed} ranks, AdaLoRA \textcolor{DarkPurple}{\textbf{prunes}} ranks, while ElaLoRA both \textcolor{DarkPurple}{\textbf{prunes}} and \textcolor{DarkOrange}{\textbf{expands}} ranks during training. $R_i$ denotes the importance score of the $i^{th}$ rank.}
    \label{fig:method_overview}
\end{figure}

To overcome these limitations, we propose ElaLoRA, a novel adaptive and dynamic LoRA framework that simultaneously prunes and expands ranks (as shown in Figure~\ref{fig:method_overview}). By dynamically reallocating computational resources to the most critical layers, ElaLoRA ensures that essential layers receive more capacity while redundant ranks are removed. ElaLoRA operates through three key components: 1) SVD-based adaptation strategy; 2) importance score calculation to quantify the significance of each rank based on loss gradients; and 3) a dynamic rank learning algorithm that reallocates ranks at scheduled intervals. Experimental results across multiple Natural Language Understanding (NLU) \citep{wang2018glue}, Natural Language Generation (NLG) \citep{narayan2018don}, and Visual Task \citep{zhai2019large} benchmarks demonstrate that ElaLoRA consistently outperforms existing PEFT methods under various parameter budgets. Notably, ElaLoRA achieves better average GLUE results with $r=2$ than other PEFT methods at $r=4$, making it particularly well-suited for resource-constrained environments. Our key contributions include:

\begin{itemize}
    \item We introduce ElaLoRA, the first method to the best of our knowledge that enables both rank pruning and expansion simultaneously during fine-tuning. Comparisons are shown in Table~\ref{tab:method_comparison}.

    \item We conduct extensive experiments across multiple benchmarks under different parameter budgets. Our results consistently demonstrate the effectiveness of ElaLoRA, outperforming existing PEFT methods in performance.
    
    \item We conduct analysis to verify that the layers and matrices identified as highly important for a specific task are indeed significant for that task, providing a principled validation of our adaptive rank allocation method.
\end{itemize}

\begin{table}[h]
    \centering
    \scriptsize 
    \centering
    \renewcommand{\arraystretch}{1.2}
    \setlength{\tabcolsep}{4pt}
    \begin{tabular}{lccc}
        \toprule[2pt]
        \textbf{Method} & \textbf{Pruning} & \textbf{Expansion} & \textbf{Dynamic Rank Allocation} \\
        \midrule[1pt]
        LoRA \citep{hu2022lora} & \No & \No & Fixed rank for all layers \\
        AdaLoRA \citep{zhang2023adalora} & \Yes & \No & Adaptive pruning via SVD \\
        SaLoRA \citep{hu2023structure} & \Yes & \No & \( L_0 \)-norm-based adaptive pruning \\
        IncreLoRA \citep{zhang2023increlora} & \No & \Yes & Heuristic-based rank expansion \\
        DoRA \citep{mao2024dora} & \Yes & \No & Decomposes into rank-one components for pruning \\
        AutoLoRA \citep{zhang2024autolora} & \Yes & \No & Meta-learning-based pruning \\
        SoRA \citep{ding2023sparse} & \Yes & \No & Gated component-wise filtering \\
        DyLoRA \citep{valipour2022dylora} & \No & \No & Rank sampling from a predefined distribution \\
        \midrule[1pt]
        \textbf{ElaLoRA (Ours)} & \Yes & \Yes & Fully dynamic rank reallocation \\
        \bottomrule[2pt]
    \end{tabular}
    \caption{Comparison of ElaLoRA with existing PEFT methods. ElaLoRA is the only approach that supports both rank pruning and expansion.}
    \label{tab:method_comparison}
\end{table}

\section{Background and Related Work}
\label{sec:background}
Fine-tuning large-scale pre-trained language models (LLMs) is an important technique for adapting them to domain-specific applications. In full model fine-tuning, all model parameters are updated during training, which is computationally expensive and memory-intensive. To address this, Adapter Tuning \citep{houlsby2019parameter, pfeiffer2020adapterfusion, he2021towards, rebuffi2017learning} introduces small trainable modules—adapter layers—inserted between transformer layers. BitFit \citep{zaken2021bitfit} adopts an even more parameter-efficient strategy by fine-tuning only the bias terms in the model. Low-Rank Adaptation (LoRA) \citep{hu2022lora} has gained popularity due to its strong performance-efficiency tradeoff. QLoRA~\citep{dettmers2023qlora} builds on this by quantizing the frozen base model to 4-bit precision, significantly reducing memory usage while maintaining performance.

However, despite its efficiency, LoRA may limit the model’s ability to memorize domain-specific knowledge and generalize to downstream tasks \citep{biderman2024lora, han2024sltrain, sui2024elrt, jiang2024mora, zhao2024galore}. A key limitation lies in its use of a fixed rank across all layers, which overlooks the fact that different layers contribute unequally to model adaptation. This uniform allocation can lead to inefficient use of trainable parameters—underfitting in layers that require more capacity and overfitting or wasted capacity in others.

To address this issue, several adaptive rank allocation methods have been developed to dynamically adjust ranks based on importance \citep{mao2025survey}. One approach is singular value decomposition (SVD)-based rank allocation, where the LoRA weight matrix is decomposed, and unimportant singular values are pruned. AdaLoRA \citep{zhang2023adalora} follows this strategy by regularizing the orthogonality of singular vectors and selectively removing less significant singular values, while SaLoRA \citep{hu2023structure} employs an \( L_0 \)-norm-based importance metric for rank selection. 

An alternative approach uses single-rank decomposition (SRD), where LoRA matrices are broken down into multiple single-rank components, allowing independent pruning. DoRA \citep{mao2024dora} decomposes LoRA matrices into rank-one components and prunes those with low importance scores. AutoLoRA \citep{zhang2024autolora} extends this method by applying meta-learning to determine rank importance, while SoRA \citep{ding2023sparse} introduces gating units to filter rank components dynamically. 

Another strategy for dynamic rank allocation is rank sampling. For example, DyLoRA \citep{valipour2022dylora} employs this approach by randomly sampling rank values at each training step and truncating LoRA matrices accordingly. Qdylora \citep{rajabzadeh2024qdylora} extends this framework to quantized models by combining dynamic rank allocation with memory-efficient quantization.

\section{Methodology}
\label{sec:methodology}
ElaLoRA continuously prunes redundant ranks while expanding ranks in layers that need additional capacity. The core components of ElaLoRA include:  
1) SVD-based Low-Rank Adaptation to enable the pruning of the least impactful ranks, 
2) Importance Score Calculation that evaluates each rank’s significance based on its impact on loss gradients,
3) Dynamic Rank Learning that prunes and reallocates ranks at scheduled intervals.  

\subsection{SVD-Based Low-Rank Adaptation}
ElaLoRA builds upon SVD-based parameterization, which has been shown to improve adaptation efficiency by allowing more precise rank adjustments \citep{zhang2023adalora}. Given a pre-trained weight matrix \( W \), we define its update as $W = W^{(0)} + \Delta = W^{(0)} + P \Lambda Q$,

where \( P \in \mathbb{R}^{d_1 \times r} \) and \( Q \in \mathbb{R}^{r \times d_2} \) are the left and right singular vectors of \( \Delta \), and \( \Lambda \in \mathbb{R}^{r \times r} \) is a diagonal matrix of singular values. The rank \( r \) is dynamically adjusted during training, ensuring \( r \ll \min(d_1, d_2) \).

To maintain numerical stability and SVD property ($P^TP=QQ^T=I$), we enforce orthogonality constraints on the singular vectors: $R(P, Q) = \|P^\top P - I\|_F^2 + \|QQ^\top - I\|_F^2.$

\subsection{Importance Score Computation}

To determine which ranks should be pruned or expanded, we compute an importance score based on the sensitivity of each weight to the loss function, following the practice of prior work on AdaLoRA \citep{zhang2023adalora}:

\begin{equation}
s(w) = |w \cdot \frac{\partial L}{\partial w}|
\end{equation}

where \( L \) is the loss function. This gradient-based importance measure uses first-order Taylor expansion to approximate the impact of removing weight and has been widely used in structured pruning approaches \citep{molchanov2019importance,liang2021super, sanh2020movement, zhang2022platon}. 

For each weight matrix, the importance of an entire rank $i$ is aggregated across its singular values and corresponding singular vectors:

\begin{equation}
S_i = s(\lambda_i) + \frac{1}{d_1} \sum_{j=1}^{d_1} s(P_{ji}) + \frac{1}{d_2} \sum_{j=1}^{d_2} s(Q_{ij})
\label{eq:rank_importance}
\end{equation}

where \( \lambda_i \) is the \( i \)-th singular value and \( P_{ji} \) and \( Q_{ij} \) are components of the left and right singular vectors.

To improve stability, we apply an exponential moving average to smooth the sensitivity \( \bar{I} \) and uncertainty \( \bar{U} \), and compute the final importance score \citep{zhang2022platon}:

\begin{equation}
\bar{I}^{(t)} = \beta_1 \bar{I}^{(t-1)} + (1 - \beta_1) I^{(t)},  \hspace{8pt} \bar{U}^{(t)} = \beta_2 \bar{U}^{(t-1)} + (1 - \beta_2) |I^{(t)} - \bar{I}^{(t)}|, s^{(t)} = \bar{I}^{(t)} \cdot \bar{U}^{(t)}
\end{equation}

where $I^{(t)}(w) = |w \cdot \frac{\partial L}{\partial w}|$ and \( 0 < \beta_1, \beta_2 < 1 \) control smoothing. This formulation ensures ranks are dynamically adjusted based on their true contribution to performance while mitigating noise from stochastic updates.

\subsection{Dynamic Rank Learning}
ElaLoRA uses a three-phase training schedule to ensure efficient rank allocation throughout fine-tuning:

\paragraph{Warm-up}  
For the first \( t_{\text{warmup}} \) iterations, ranks remain fixed, allowing the model to initialize its representations before modifications.

\paragraph{Dynamic Rank Adjustment}  
Ranks are adjusted every \( t_{\text{adjust}} \) iterations through the following steps:
\begin{enumerate}
    \item Identify the \( k \) least important ranks in each weight matrix. In each rank adjustment phase, each weight matrix can be pruned and expanded by at most $k$ ranks.
    \item Sort these \( k \times N \) ranks across all matrices by importance score.
    \item Prune \( b \) of the least important ranks, selecting the lowest-ranked entries across all matrices. This ensures that ranks with minimal contribution to model performance are removed first.
    \item Expand ranks in matrices where the least important retained ranks have relatively high importance scores, indicating a greater need for capacity. This prioritizes matrices where even the lowest-ranked components play a significant role, ensuring effective resource allocation. We expand \( b \) ranks in total.
\end{enumerate}
This ensures expressivity is preserved while avoiding computational redundancy.

\paragraph{Stabilization}  
For the final \( t_{\text{stabilize}} \) iterations, rank updates are frozen, allowing the model to converge smoothly.

The full ElaLoRA training procedure is outlined in Algorithm~\ref{alg:ElaLoRA}.

\begin{algorithm}
\caption{ElaLoRA}
\label{alg:ElaLoRA}
\begin{algorithmic}[1]
\Require $I$: total iterations, $T$: rank adjustment schedule, $r$: initial average rank
\Require $k$: max ranks pruned/expanded per weight matrix per step
\Require $b^{(0)}$: total ranks pruned/expanded per step
\Require Pre-trained model weights $W^{(0)}$ 

\State \textbf{Initialize:} $(P_{0:N-1}, \Lambda_{0:N-1}, Q_{0:N-1})$ for all weight matrices with rank $r$

\For{$i = 1$ to $I$}
    \State \Call{UpdateWeights}{$P, \Lambda, Q$} 
    \State $S_{0:N-1} \gets$ \Call{CalculateImportanceScores}{$P, Q, \Lambda, \nabla$}
     
    \If{$i \in T$} 
        \State $C \gets$ \Call{LeastImportantRanks}{$S, k$}  \hfill \textcolor{gray}{\textit{▹ Identify top-$k$ least important ranks per matrix}}
        
        \State \Call{Sort}{$C$}  \hfill \textcolor{gray}{\textit{▹ Sort ranks in ascending order of importance}}
        
        \State \Call{PruneRanks}{$C[:b^{(i)}], P, \Lambda, Q$}  \hfill \textcolor{gray}{\textit{▹ Remove the least important $b$ ranks}}
        
        \State \Call{ExpandRanks}{$C[-b^{(i)}:], P, \Lambda, Q$} 
        \Statex \hspace{1.2em} \hfill \textcolor{gray}{\textit{▹ Expand most important weight matrices with Gram-Schmidt initialization}}
    \EndIf
\EndFor

\State \Return Fine-tuned parameters $(P, \Lambda, Q)$

\end{algorithmic}
\end{algorithm}

\subsection{Dynamic Rank Scheduler}
To dynamically adjust the number of ranks pruned and expanded (\( b \) in Algorithm~\ref{alg:ElaLoRA}) during training, we introduce a dynamic rank scheduler. This scheduler gradually modulates the rank adjustment aggressiveness, ensuring a smooth transition from the initial adaptation phase to the final convergence phase, thereby improving rank search stability. We demonstrate the effectiveness of the scheduler in Section \ref{sec:ablation}.

Specifically, the scheduler regulates the total number of ranks pruned and expanded at each adjustment interval, adapting to the training progress. The scheduling function is defined as:

\begin{equation}
P = \frac{\text{current\_step} - \text{initial\_warmup}}{\text{total\_step} - \text{final\_stabilization} - \text{initial\_warmup}}
\end{equation}

To provide a gradual reduction in the number of rank adjustments, we define using a cubic polynomial interpolation as follows:

\begin{equation}
b^{(t)} = \text{round}\left( b \times (1-P^3) \right)
\end{equation}
where we start with $b$ and use $b^{(t)}$ for the $t^{th}$ rank adjustment step in Algorithm \ref{alg:ElaLoRA}.

\section{Experiments}
\label{sec:experiments}
\subsection{Setup}
We implemented ElaLoRA for fine-tuning DeBERTaV3-base \citep{he2021debertav3} and BART-base \citep{lewis2019bart} on natural language understanding tasks (NLU) using the GLUE benchmark datasets \citep{wang2018glue} and the natural language generation tasks (NLG) using the XSum datasets \citep{narayan2018don}. Additionally, to demonstrate the versatility of our approach, we extended ElaLoRA to the vision domain by applying it to a ViT-B/16 \citep{dosovitskiy2021an} model, evaluating on a subset of VTAB tasks \citep{zhai2019large}.

Our analysis mainly focuses on comparing ElaLoRA's performance to fixed-rank LoRA \citep{hu2022lora} and pruning-only AdaLoRA \citep{zhang2023adalora} under varying configurations. Additionally, we reference performance results for full model fine-tuning, Adapter Tuning \citep{houlsby2019parameter,pfeiffer2020adapterfusion}, BitFit \citep{zaken2021bitfit}, and DoRA \citep{mao2024dora} from existing literature to provide a comprehensive comparative analysis.

The experiments were conducted on NVIDIA A100 GPUs with 80GB of memory, using the PyTorch framework \citep{paszke2019pytorch}. ElaLoRA was implemented as a modular extension to the LoRA framework \citep{hu2022lora}, with support for rank pruning and expansion based on the calculated importance scores. Additional tools included the Hugging Face Transformers library for dataset preprocessing and model initialization \citep{wolf2019huggingface}.

\subsection{Natural Language Understanding}

\paragraph{Datasets.}
We conducted experiments on the GLUE benchmark datasets \citep{wang2018glue}. GLUE includes a variety of tasks such as sentence relationship
recognition, sentiment analysis, and natural language reasoning. For example, MRPC is a binary classification task to determine whether two sentences are paraphrases of each other, and RTE is a binary classification task to determine if a hypothesis is entailed by a premise. MNLI, SST-2, QNLI, PRTE, MRPC, and QQP use accuracy, while we also report F1 score of QQP; CoLA uses Matthews correlation coefficient (MCC); and STS-B uses Pearson correlation coefficients. Dataset details are summarized in Appendix~\ref{app:glue}. The experiments were performed using the DeBERTa-v3-base model, a transformer architecture with 183 million parameters \citep{he2021debertav3}. 

\paragraph{Implementation Details.}  
We compare our method primarily with LoRA \citep{hu2022lora} and AdaLoRA \citep{zhang2023adalora}, evaluating performance across three different rank settings: 2, 4, and 10.  
Since AdaLoRA initializes with ranks 1.5 times the final ranks, its parameter count is 50\% higher than that of LoRA and ElaLoRA at the same target rank.  

To ensure consistency and clarity, we adopt intuitive experimental settings, such as using epoch counts that are multiples of 5 across all tasks. Details are summarized in Appendix ~\ref{app:nlu_config}.
Additionally, we cite performance results from AdaLoRA \citep{zhang2023adalora}, DoRA \citep{mao2024dora}, and DyLoRA \citep{valipour2022dylora} to provide a comprehensive comparative analysis. Our experiments were conducted with fewer training epochs across all tasks compared to the cited results, highlighting ElaLoRA’s efficiency in achieving competitive performance with reduced computational cost.

\paragraph{Main Results.}
Table~\ref{tab:nlu_results} summarizes the results on the GLUE tasks. ElaLoRA consistently achieves the highest average performance across all three tested rank and parameter budget levels. Notably, ElaLoRA with rank 2 outperforms AdaLoRA \citep{zhang2023adalora} with rank 4 and all other methods, including LoRA \citep{hu2022lora}, even at rank 10. Specifically, ElaLoRA is more effective when using higher ranks, potentially due to a broader search space for optimal ranks. For example, when $r=10$, ElaLoRA performs the best in 7 out of the 8 tasks in GLUE \citep{wang2018glue}. 

\begin{table*}[h]
    
    \tiny
    \centering
    \renewcommand{\arraystretch}{1.2}
    \setlength{\tabcolsep}{4pt}
    \begin{tabular}{lccc c c c c c c c c}
    \toprule[1pt]
    \textbf{Method} & \textbf{\# Params} &\textbf{Rank} & \textbf{MNLI} & \textbf{SST-2} & \textbf{CoLA} & \textbf{QQP} & \textbf{QNLI} & \textbf{RTE} & \textbf{MRPC} & \textbf{STS-B} & \textbf{All} \\
        & & & \textbf{Acc.} & \textbf{Acc.} & \textbf{Mcc} & \textbf{Acc./F1} & \textbf{Acc.} & \textbf{Acc.} & \textbf{Acc.} & \textbf{Corr.} & \textbf{Avg.} \\

        \midrule[1pt]
        Full FT* & 184M & & 89.90 & 95.63 & 69.19 & 92.40/89.80 & 94.03 & 83.75 & 89.46 & 91.60 & 88.09 \\
        \hline
        BitFit* & 0.1M & & 89.37 & 94.84 & 66.96 & 88.41/84.95 & 92.24 & 78.70 & 87.75 & 91.35 & 86.02 \\
        \hline
        H-Adapter* & 0.31M & & 86.31 & 93.54 & 61.76 & 90.16 & 92.52 & 78.56 & 88.64 & 90.88 & 85.30 \\
        P-Adapter* & 0.30M & & 86.23 & 93.24 & 62.92 & 89.94 & 92.59 & 79.07 & 88.74 & 90.44 & 85.40 \\
        DoRA* & 0.34M & & 87.45 & 91.28 & 64.90 & 90.64 & 92.93 & 79.15 & 89.72 & 91.28 & 86.38 \\
        DyLoRA* & & $r=2$ & 86.02 & 93.81 & 59.91 & 89.33 & 92.39 & 76.03 & 91.66 & 90.60 & 84.97 \\
        \hdashline
        LoRA &  0.33M & & 88.43 & 94.49 & \textbf{69.77} & 90.92/88.02 & 93.84 & 83.03 & 88.48 & 91.13 & 87.51 \\
        AdaLoRA & 0.49M & $r=2$ & 89.05 & 94.95 & 66.87 & 90.99/88.11 & \textbf{94.33} & 84.84 & 86.51 & \textbf{91.60} & 87.39 \\
        ElaLoRA & 0.33M &  & \textbf{89.32} & \textbf{95.53} & 67.38 & \textbf{91.21/88.33} & 93.70 & \textbf{85.92} & \textbf{90.44} & 90.61 & \textbf{88.01} \\

        \hline
        H-Adapter* & 1.20M & & 86.53 & 93.73 & 62.62 & 90.83 & 92.82 & 80.43 & 89.90 & 90.16 & 85.88 \\
        P-Adapter* & 1.19M & & 86.75 & 93.83 & 63.87 & 90.53 & 92.61 & 80.51 & 89.51 & 90.65 & 86.03 \\
        DoRA* & 1.31M & & 87.81 & 91.34 & 65.35 & \textbf{91.32} & 92.97 & 81.73 & 90.05 & 91.34 & 86.97 \\
        DyLoRA* & & $r=4$ & 86.82 & 94.40 & 59.81 & 89.80 & 92.91 & 77.40 & 92.06 & 90.86 & 85.53 \\
        \hdashline
        LoRA & 0.66M & & \textbf{89.46} & 94.15 & 67.15 & 91.26/88.50 & 93.84 & 85.55 & 88.48 & 91.17 & 87.63 \\
        AdaLoRA & 0.99M & $r=4$ & 88.92 & 95.53 & 67.70 & 91.10/88.25 & 94.12 & 86.28 & 87.74 & \textbf{91.74} & 87.89 \\
        ElaLoRA & 0.66M & & 89.44 & \textbf{95.64} & \textbf{70.15} & \textbf{91.26/88.60} & \textbf{94.44} & \textbf{87.36} & \textbf{90.44} & 91.10 & \textbf{88.72} \\

        \hline
        LoRA & 1.66M & & 89.11 & 92.31 & 67.06 & 91.34/88.62 & 94.08 & 87.72 & 88.97 & 91.40 & 87.75 \\
        AdaLoRA & 2.49M & $r=10$ & 89.27 & 95.76 & 70.02 & 91.16/88.37 & 94.31 & 87.73 & 88.72 & \textbf{91.67} & 88.58 \\
        \midrule[0.8pt]
        \textbf{ElaLoRA} & 1.66M & & \textbf{89.51} & \textbf{95.87} & \textbf{70.19} & \textbf{91.42/ 88.83} & \textbf{94.36} & \textbf{88.81} & \textbf{88.98 }& 91.59 & \textbf{88.84} \\
        \bottomrule[1pt]
        
    \end{tabular}
    \vspace{10pt}
  \caption{Performance comparison of LoRA, AdaLoRA, ElaLoRA, and other fine-tuning methods across different ranks and parameter budgets.  
Methods marked with * are results cited from the AdaLoRA \citep{zhang2023adalora}, DoRA \citep{mao2024dora}, and DyLoRA \citep{valipour2022dylora} papers.  
For QQP, both accuracy and F1 scores are reported when available, though some methods lack F1 scores. Similarly, we provide both parameter counts and rank information for LoRA, AdaLoRA, and ElaLoRA, whereas this information is unavailable for certain other methods.}

    \label{tab:nlu_results}
\end{table*}

\subsection{Natural Language Generation}
\paragraph{Datasets.}
To benchmark ElaLoRA against state-of-the-art methods in natural language generation (NLG), we fine-tune a BART-large model \citep{lewis2019bart} and evaluate its performance on the XSum dataset \citep{narayan2018don}. We report ROUGE-1/2/L metrics.

\paragraph{Implementation Details.}
Similar to our evaluation in Natural Language Understanding (NLU), we compare ElaLoRA with LoRA \citep{hu2022lora} and AdaLoRA \citep{zhang2023adalora}, fine-tuning all three methods under the same experimental setup. Performance is assessed at two rank settings: 2 and 6. We use a beam length of 8 and a batch size of 64 for decoding. For a detailed configuration, please refer to Appendix \ref{app:nlg_config}.

\paragraph{Main Results.}
\begin{wraptable}{r}{0.55\textwidth}
    \small
    \vspace{-5mm}
    \begin{tabular}{lccccc}
        \toprule[2pt]
        \textbf{Method} & \textbf{Rank} & \textbf{\# Params} & \textbf{XSum} \\
        \midrule[1pt]
        Full FT & - & 124.65M & 40.61/17.76/32.91 \\
        \midrule[1pt]
        LoRA & $r=2$ & 0.41M & 36.61/14.28/29.23 \\
        AdaLoRA & $r=2$ & 0.41M & 36.97/14.42/29.42 \\
        ElaLoRA & $r=2$ & 0.41M & \textbf{37.24/14.66/29.76} \\
        \midrule[1pt]
        LoRA & $r=6$ & 1.22M & 37.64/15.30/30.19 \\
        AdaLoRA & $r=6$ & 1.22M & 37.80/15.18/30.28 \\
        \midrule[0.8pt]
        \textbf{ElaLoRA} & $r=6$ & 1.22M & \textbf{38.00/15.30/30.42} \\
        \bottomrule[2pt]
    \end{tabular}
    \captionof{table}{Results of fine-tuning Bart-base on XSum. The best results are in bold.}
    \label{tab:xsum_results}
\end{wraptable}

Table~\ref{tab:xsum_results} presents the results of fine-tuning BART on XSum. We also cited Full FT result from DoRA \citep{mao2024dora}. Notice that ElaLoRA consistently outperforms LoRA and AdaLoRA across both rank settings. For example, at \( r=6 \), ElaLoRA further improves to 38.00/15.30/30.42, achieving the best performance among these three methods.

\subsection{Visual Task}

\paragraph{Datasets.}
We conduct experiments on a subset of VTAB-1k \citep{zhai2019large}, which evaluates performance across diverse visual tasks. Each dataset contains 1000 images (800 for training, 200 for evaluation) spanning three categories: natural, specialized, and structured. Natural tasks involve image classification and object recognition; specialized tasks target fine-grained, domain-specific images; structured tasks assess spatial and relational reasoning. We randomly select two datasets per category and report classification accuracy. Dataset details are provided in Appendix~\ref{app:vtab_dataset}.

\paragraph{Implementation Details.}  
The base model employed in our experiments is ViT-B/16, pre-trained on ImageNet-22K. All methods are evaluated under a consistent experimental setup, with all experiments independently conducted by our team. We fine-tune the model for 100 epochs on each dataset using a batch size of 16 and a rank of 8. For detailed configuration settings, please refer to Appendix~\ref{app:vit_config}.

\paragraph{Main Results.}  
Table~\ref{tab:vit_results} shows that ElaLoRA outperforms both LoRA and AdaLoRA in most tasks. Specifically, ElaLoRA achieves higher accuracy in natural tasks (CIFAR-100: 61.25, SVHN: 74.60) and specialized tasks (Eurosat: 94.26, Resisc45: 80.71), while also providing competitive performance in structured tasks. The aggregated average of 64.88 underscores the effectiveness of our approach in balancing diverse visual challenges.

\begin{table*}[ht]
\tiny
\centering
\renewcommand{\arraystretch}{1.2}
\begin{tabular}{lcc|cc|cc|cc|c}
\toprule
\textbf{Method} & \textbf{\# Params} & \textbf{Rank} & 
\multicolumn{2}{c|}{\textbf{Natural}} & 
\multicolumn{2}{c|}{\textbf{Specialized}} & 
\multicolumn{2}{c|}{\textbf{Structured}} & 
\textbf{Average} \\
\cmidrule(lr){4-5} \cmidrule(lr){6-7} \cmidrule(lr){8-9}
 &  &  & CIFAR-100 & SVHN & Eurosat & Resisc45 & dSpr\-Ori & DMLab &  \\
\midrule
LoRA & 1.19M & $r=8$ & 53.75 & 73.83 & 93.85 & 79.63 & \textbf{40.62} & \textbf{41.5}& 63.86\\
AdaLoRA    & 1.48M & $r=8$ & 53.90 & 72.30 & 93.22 & 79.22 & 40.54 &39.62 & 63.13\\
\midrule[0.8pt]
\textbf{ElaLoRA}     & 1.19M & $r=8$ & \textbf{61.25} & \textbf{74.60} & \textbf{94.26} & \textbf{80.71} & 39.2 &39.30  & \textbf{64.88}\\
\bottomrule
\end{tabular}
\caption{Results of fine-tuning ViT-B/16 on VTAB-1k tasks. The best results are in bold.}
\label{tab:vit_results}
\end{table*}

\subsection{Final Rank Distribution Analysis}
\begin{figure}[h!]
    \centering
    \begin{minipage}{0.49\textwidth}
        \centering
        \includegraphics[width=\linewidth]{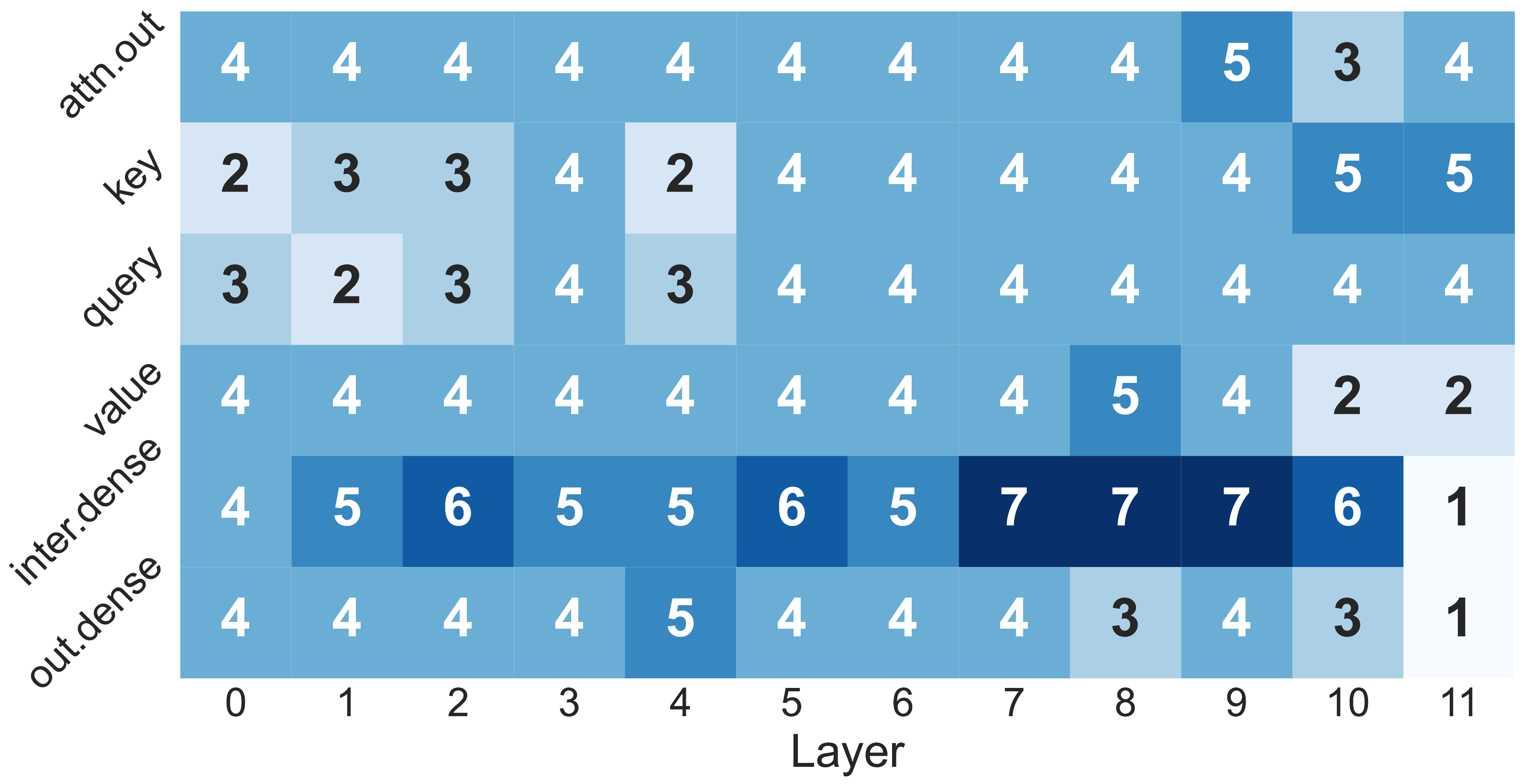}
        \caption{RTE Final Rank Heatmap $(r=4)$}
        \label{fig:rte_r4}
    \end{minipage}
    \hfill
    \begin{minipage}{0.49\textwidth}
        \centering
        \includegraphics[width=\linewidth]{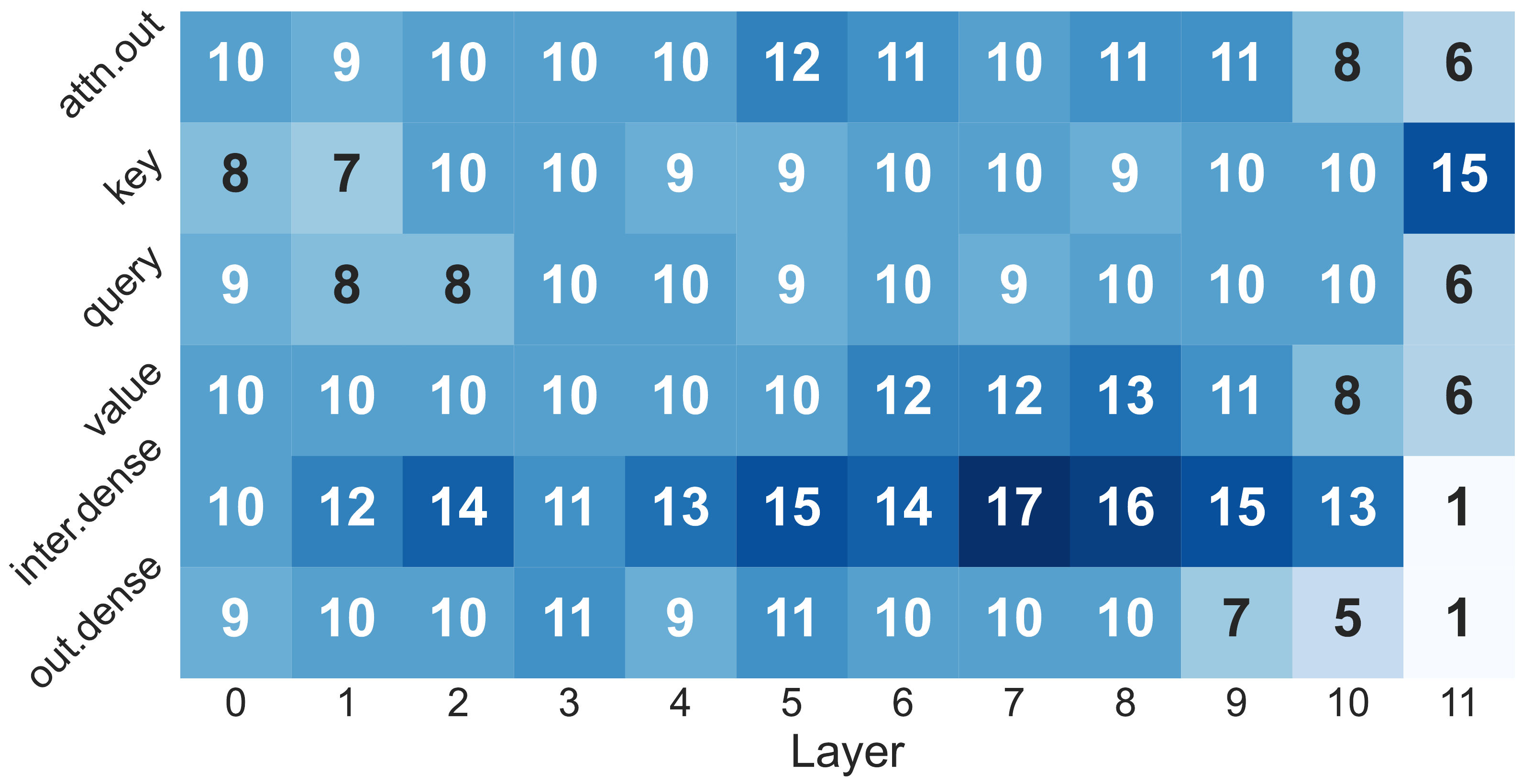}
        \caption{RTE Final Rank Heatmap $(r=10)$}
        \label{fig:rte_r10}
    \end{minipage}
\end{figure}

We plot ElaLoRA's final rank distributions to show that layers with higher ranks indeed contribute more to performance. Figure~\ref{fig:rte_r4} and Figure~\ref{fig:rte_r10} illustrate the final rank distributions by ElaLoRA on the RTE task with \( r=4 \) and \( r=10 \). In both cases, the intermediate feed-forward layers receive the highest rank allocations, whereas the final projection output layers are assigned significantly fewer ranks. Also notice that the highest ranks ElaLoRA reached are 7 ($r=4$) and 17 ($r=10$), which is higher than what AdaLoRA \citep{zhang2023adalora} can reach if they start with 1.5 times final ranks.
To validate ElaLoRA’s rank allocation strategy, we conducted three additional experiments with different rank distributions: (1) excluding ranks from all intermediate feed-forward layers, (2) excluding ranks from the final projection layer, and (3) assigning ranks only to the intermediate feed-forward layers.

\begin{wrapfigure}{r}{0.52\textwidth}
    \centering
    \vspace{-10mm}
    \includegraphics[width=0.5\textwidth]{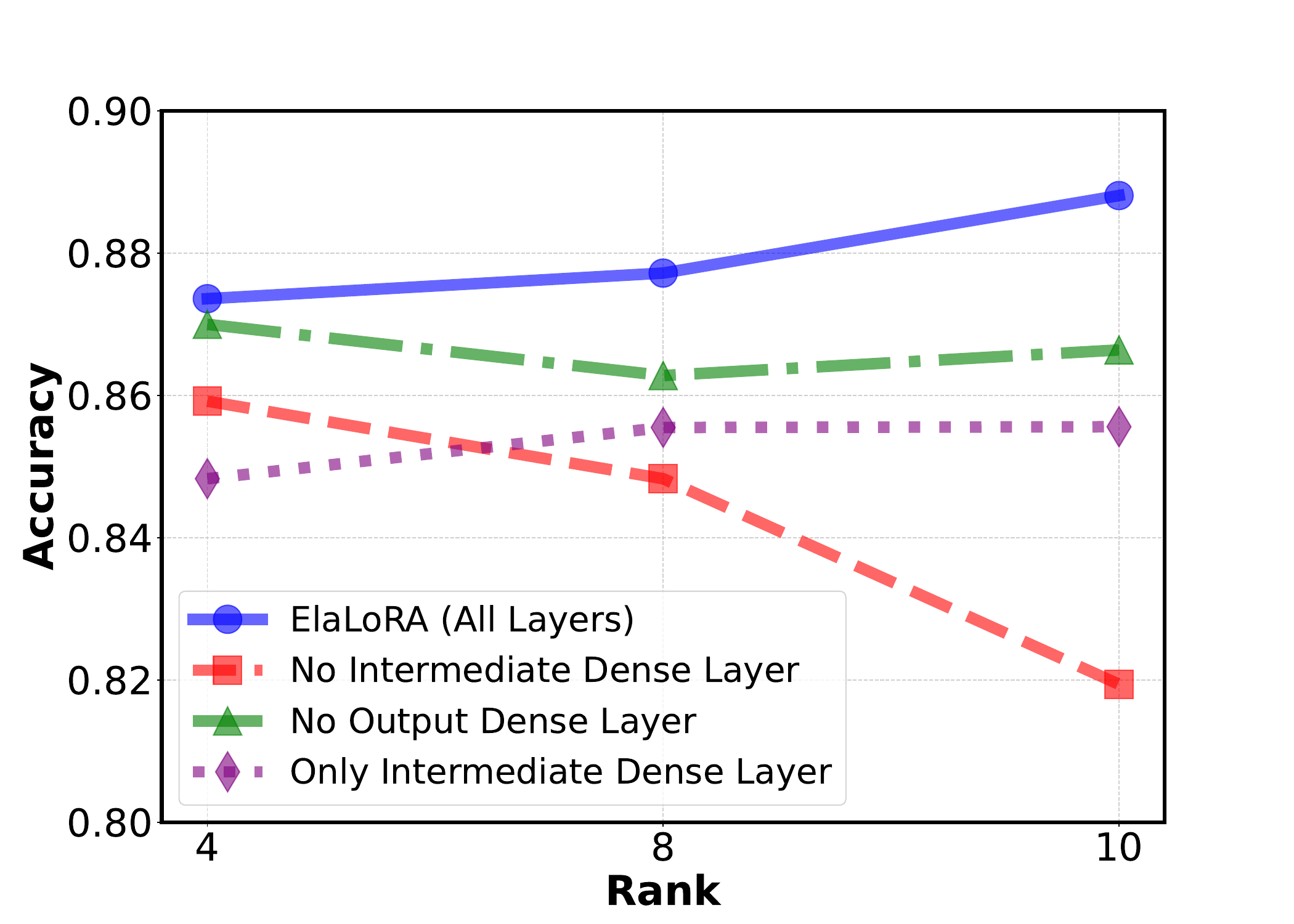}
    \vspace{-2mm}
    \captionof{figure}{RTE Performance Comparisons}
    \label{fig:rte_plot}
\end{wrapfigure}

As shown in Figure~\ref{fig:rte_plot}, removing the intermediate feed-forward layers during fine-tuning leads to a significant drop in final performance, whereas removing the final projection layer has a relatively minor effect. Furthermore, allocating ranks exclusively to the intermediate feed-forward layers yields competitive performance, highlighting their crucial role in task adaptation. We have provided more analysis and final rank heatmaps for other GLUE tasks \citep{wang2018glue} in Appendix \ref{app:heatmap}.

\subsection{Importance Score Analysis}
\begin{figure}[h!]
    \centering
    \begin{minipage}{0.46\textwidth}
        \centering
        \includegraphics[width=\linewidth]{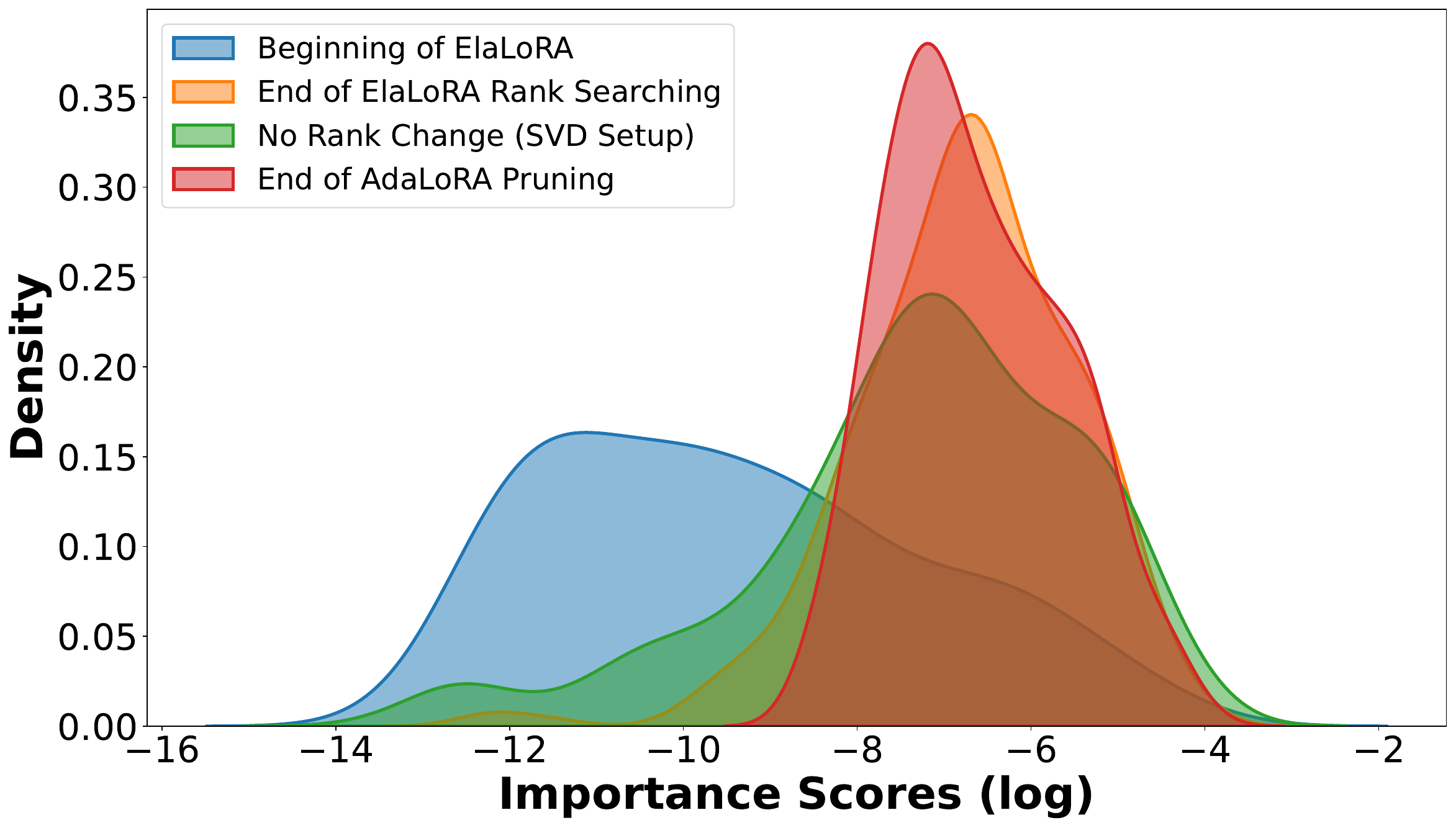}
    \end{minipage}
    \hfill
    \begin{minipage}{0.46\textwidth}
        \centering
        \includegraphics[width=\linewidth]{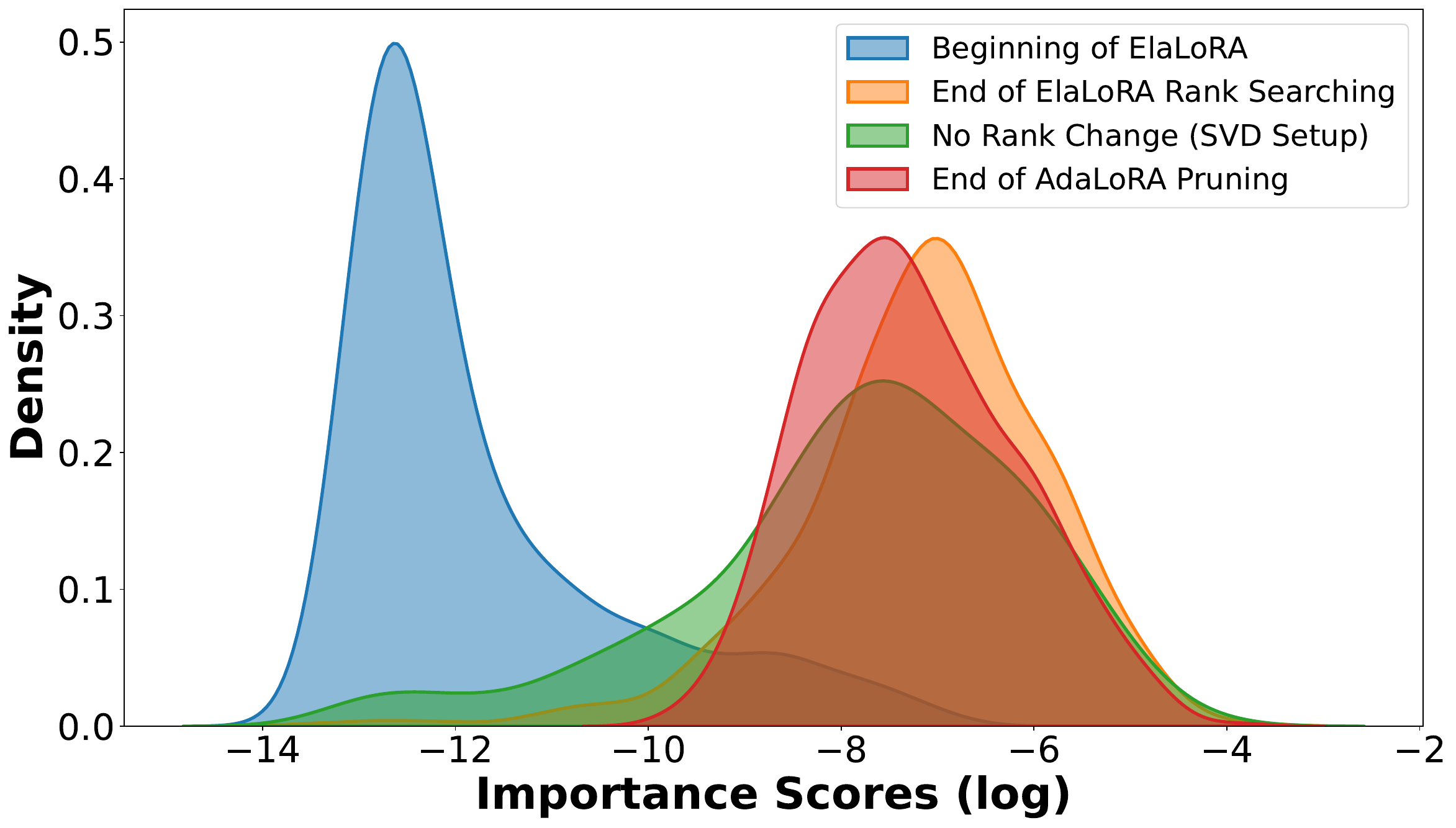}
    \end{minipage}
    \caption{Comparison of importance score distributions for the MRPC task at $r=4$ (left) and $r=10$ (right) settings with different methods. \textcolor{cyan}{Beginning of ElaLoRA}: Right after the \textit{Warm-up Phase}, where all ranks are still fixed. \textcolor{orange}{End of ElaLoRA Rank Searching}: Right after the \textit{Dynamic Rank Adjustment Phase}, where ElaLoRA has finished dynamically allocating ranks. \textcolor{DarkGreen}{No Rank Change (SVD Setup)}: Trained for the same number of iterations as ElaLoRA but without any rank updates (i.e., ranks remain static). \textcolor{red}{End of AdaLoRA Pruning}: Trained for the same number of iterations as ElaLoRA, but using AdaLoRA's pruning-based approach.  }
    \label{fig:mrpc_distribution}
\end{figure}

To analyze how ElaLoRA optimizes rank selection, we present the importance score distributions for the MRPC task at ranks \( r=4 \) and \( r=10 \) in Figure~\ref{fig:mrpc_distribution}. The importance scores are computed at different stages of training using different methods. Notice that:
\begin{itemize}
\item Both \textcolor{orange}{ElaLoRA} and \textcolor{red}{AdaLoRA} successfully remove unimportant ranks, as indicated by the leftward removal of density mass compared to the \textcolor{DarkGreen}{fixed-rank setup}.

\item At \( r=4 \), the peak of \textcolor{orange}{ElaLoRA's} importance score distribution is shifted further right than \textcolor{red}{AdaLoRA's}, while at \( r=10 \), \textcolor{orange}{ElaLoRA} shifts the entire importance score distribution to the right. It means \textcolor{orange}{ElaLoRA} fine-tuned parameters have a greater overall impact on the loss function than both \textcolor{red}{AdaLoRA} and the \textcolor{DarkGreen}{fixed-rank setup}.
\end{itemize}

\subsection{Ablation Study on the Dynamic Rank Scheduler}
\label{sec:ablation}

\begin{wraptable}{r}{0.56\textwidth}
    \small
    \vspace{-4mm}
    \centering
    \begin{tabular}{lcc}
        \toprule[1.5pt]
        \textbf{Task} & \textbf{Scheduler} & \textbf{Accuracy / ROUGE} \\
        \midrule[1pt]
        \multicolumn{3}{c}{\textit{Classification (Accuracy)}} \\
        RTE (r=2/4/10) & \Yes & \textbf{85.92} / \textbf{87.36} / \textbf{88.81} \\
                       & \No  & 84.84 / 86.60 / 87.36 \\
        MRPC (r=2/4/10) & \Yes & \textbf{90.44} / \textbf{90.44} / \textbf{88.98} \\
                        & \No  & 88.97 / 88.23 / 88.72 \\
        \midrule[0.8pt]
        \multicolumn{3}{c}{\textit{Generation (ROUGE-1/2/L)}} \\
        XSum (r=2) & \Yes & \textbf{37.24} / \textbf{14.66} / \textbf{29.76} \\
                   & \No  & 37.22 / 14.62 / 29.61 \\
        XSum (r=6) & \Yes & \textbf{38.00} / \textbf{15.30} / 30.42 \\
                   & \No  & 37.57 / 15.06 / \textbf{30.91} \\
        \bottomrule[1.5pt]
    \end{tabular}
    \caption{Dynamic Rank Scheduler Analysis}
    \label{tab:scheduler_ablation}
\end{wraptable}

To evaluate the impact of the dynamic rank scheduler, we compare ElaLoRA’s performance with and without it across classification (RTE, MRPC) and generation (XSum) tasks. We report accuracy for classification and ROUGE-1/2/L for generation. As shown in Table~\ref{tab:scheduler_ablation}, the scheduler consistently improves accuracy across RTE and MRPC at all ranks and enhances ROUGE scores on XSum. These results demonstrate that the rank adjustment scheduler does improve ElaLoRA performance.

\section{Conclusion}
\label{sec:conclusion}
In this work, we introduced \textbf{ElaLoRA}, a novel parameter-efficient fine-tuning (PEFT) method that dynamically prunes and expands ranks based on importance scores, ensuring that the most impactful layers receive additional capacity while removing redundant ranks. This adaptive rank learning mechanism enables more efficient model adaptation across diverse NLP and Vision tasks.

Our empirical results demonstrate that ElaLoRA outperforms other state-of-the-art  methods, achieving superior accuracy across multiple benchmark datasets while maintaining a lower or comparable parameter budget. Beyond performance improvements, our analysis of final rank distributions and importance score distributions confirms that ElaLoRA's rank allocation decisions align with the layers that contribute most to task-specific learning. 

\section*{Ethics Statement}

ElaLoRA enhances the efficiency of fine-tuning large language models, reducing computational costs and environmental impact. However, it inherits biases from pre-trained models and does not inherently mitigate them. We encourage fairness-aware evaluation when applying ElaLoRA in sensitive domains. Additionally, while our method improves accessibility, it can be misused for generating misinformation or biased content. We advocate for responsible deployment, dataset transparency, and adherence to ethical AI guidelines. Our work supports open research, but computational accessibility remains a challenge that requires broader community efforts.

\newpage

\bibliography{colm2025_conference}

\begin{thebibliography}{44}
\providecommand{\natexlab}[1]{#1}
\providecommand{\url}[1]{\texttt{#1}}
\expandafter\ifx\csname urlstyle\endcsname\relax
  \providecommand{\doi}[1]{doi: #1}\else
  \providecommand{\doi}{doi: \begingroup \urlstyle{rm}\Url}\fi

\bibitem[Achiam et~al.(2023)Achiam, Adler, Agarwal, Ahmad, Akkaya, Aleman, Almeida, Altenschmidt, Altman, Anadkat, et~al.]{achiam2023gpt}
Josh Achiam, Steven Adler, Sandhini Agarwal, Lama Ahmad, Ilge Akkaya, Florencia~Leoni Aleman, Diogo Almeida, Janko Altenschmidt, Sam Altman, Shyamal Anadkat, et~al.
\newblock Gpt-4 technical report.
\newblock \emph{arXiv preprint arXiv:2303.08774}, 2023.

\bibitem[Bi et~al.(2024)Bi, Chen, Chen, Chen, Dai, Deng, Ding, Dong, Du, Fu, et~al.]{bi2024deepseek}
Xiao Bi, Deli Chen, Guanting Chen, Shanhuang Chen, Damai Dai, Chengqi Deng, Honghui Ding, Kai Dong, Qiushi Du, Zhe Fu, et~al.
\newblock Deepseek llm: Scaling open-source language models with longtermism.
\newblock \emph{arXiv preprint arXiv:2401.02954}, 2024.

\bibitem[Biderman et~al.(2024)Biderman, Portes, Ortiz, Paul, Greengard, Jennings, King, Havens, Chiley, Frankle, et~al.]{biderman2024lora}
Dan Biderman, Jacob Portes, Jose Javier~Gonzalez Ortiz, Mansheej Paul, Philip Greengard, Connor Jennings, Daniel King, Sam Havens, Vitaliy Chiley, Jonathan Frankle, et~al.
\newblock Lora learns less and forgets less.
\newblock \emph{Transactions on Machine Learning Research}, 2024.

\bibitem[Brown et~al.(2020)Brown, Mann, Ryder, Subbiah, Kaplan, Dhariwal, Neelakantan, Shyam, Sastry, Askell, et~al.]{brown2020language}
Tom Brown, Benjamin Mann, Nick Ryder, Melanie Subbiah, Jared~D Kaplan, Prafulla Dhariwal, Arvind Neelakantan, Pranav Shyam, Girish Sastry, Amanda Askell, et~al.
\newblock Language models are few-shot learners.
\newblock \emph{Advances in neural information processing systems}, 33:\penalty0 1877--1901, 2020.

\bibitem[Dettmers et~al.(2023)Dettmers, Pagnoni, Holtzman, and Zettlemoyer]{dettmers2023qlora}
Tim Dettmers, Artidoro Pagnoni, Ari Holtzman, and Luke Zettlemoyer.
\newblock Qlora: Efficient finetuning of quantized llms.
\newblock \emph{Advances in neural information processing systems}, 36:\penalty0 10088--10115, 2023.

\bibitem[Devlin et~al.(2019)Devlin, Chang, Lee, and Toutanova]{devlin2019bert}
Jacob Devlin, Ming-Wei Chang, Kenton Lee, and Kristina Toutanova.
\newblock Bert: Pre-training of deep bidirectional transformers for language understanding.
\newblock In \emph{Proceedings of the 2019 conference of the North American chapter of the association for computational linguistics: human language technologies, volume 1 (long and short papers)}, pp.\  4171--4186, 2019.

\bibitem[Ding et~al.(2023{\natexlab{a}})Ding, Lv, Wang, Chen, Zhou, Liu, and Sun]{ding2023sparse}
Ning Ding, Xingtai Lv, Qiaosen Wang, Yulin Chen, Bowen Zhou, Zhiyuan Liu, and Maosong Sun.
\newblock Sparse low-rank adaptation of pre-trained language models.
\newblock \emph{arXiv preprint arXiv:2311.11696}, 2023{\natexlab{a}}.

\bibitem[Ding et~al.(2023{\natexlab{b}})Ding, Qin, Yang, Wei, Yang, Su, Hu, Chen, Chan, Chen, et~al.]{ding2023parameter}
Ning Ding, Yujia Qin, Guang Yang, Fuchao Wei, Zonghan Yang, Yusheng Su, Shengding Hu, Yulin Chen, Chi-Min Chan, Weize Chen, et~al.
\newblock Parameter-efficient fine-tuning of large-scale pre-trained language models.
\newblock \emph{Nature Machine Intelligence}, 5\penalty0 (3):\penalty0 220--235, 2023{\natexlab{b}}.

\bibitem[Dosovitskiy et~al.(2021)Dosovitskiy, Beyer, Kolesnikov, Weissenborn, Zhai, Unterthiner, Dehghani, Minderer, Heigold, Gelly, Uszkoreit, and Houlsby]{dosovitskiy2021an}
Alexey Dosovitskiy, Lucas Beyer, Alexander Kolesnikov, Dirk Weissenborn, Xiaohua Zhai, Thomas Unterthiner, Mostafa Dehghani, Matthias Minderer, Georg Heigold, Sylvain Gelly, Jakob Uszkoreit, and Neil Houlsby.
\newblock An image is worth 16x16 words: Transformers for image recognition at scale.
\newblock \emph{arXiv preprint arXiv:2010.11929}, 2021.

\bibitem[Gunter et~al.(2024)Gunter, Wang, Wang, Pang, Narayanan, Zhang, Zhang, Chen, Chiu, Qiu, et~al.]{gunter2024apple}
Tom Gunter, Zirui Wang, Chong Wang, Ruoming Pang, Andy Narayanan, Aonan Zhang, Bowen Zhang, Chen Chen, Chung-Cheng Chiu, David Qiu, et~al.
\newblock Apple intelligence foundation language models.
\newblock \emph{arXiv preprint arXiv:2407.21075}, 2024.

\bibitem[Han et~al.(2024)Han, Li, Huang, Hong, Takeda, Jawanpuria, and Mishra]{han2024sltrain}
Andi Han, Jiaxiang Li, Wei Huang, Mingyi Hong, Akiko Takeda, Pratik Jawanpuria, and Bamdev Mishra.
\newblock Sltrain: a sparse plus low-rank approach for parameter and memory efficient pretraining.
\newblock \emph{arXiv preprint arXiv:2406.02214}, 2024.

\bibitem[He et~al.(2021{\natexlab{a}})He, Zhou, Ma, Berg-Kirkpatrick, and Neubig]{he2021towards}
Junxian He, Chunting Zhou, Xuezhe Ma, Taylor Berg-Kirkpatrick, and Graham Neubig.
\newblock Towards a unified view of parameter-efficient transfer learning.
\newblock \emph{arXiv preprint arXiv:2110.04366}, 2021{\natexlab{a}}.

\bibitem[He et~al.(2020)He, Liu, Gao, and Chen]{he2020deberta}
Pengcheng He, Xiaodong Liu, Jianfeng Gao, and Weizhu Chen.
\newblock Deberta: Decoding-enhanced bert with disentangled attention.
\newblock \emph{arXiv preprint arXiv:2006.03654}, 2020.

\bibitem[He et~al.(2021{\natexlab{b}})He, Gao, and Chen]{he2021debertav3}
Pengcheng He, Jianfeng Gao, and Weizhu Chen.
\newblock Debertav3: Improving deberta using electra-style pre-training with gradient-disentangled embedding sharing.
\newblock \emph{arXiv preprint arXiv:2111.09543}, 2021{\natexlab{b}}.

\bibitem[Houlsby et~al.(2019)Houlsby, Giurgiu, Jastrzebski, Morrone, De~Laroussilhe, Gesmundo, Attariyan, and Gelly]{houlsby2019parameter}
Neil Houlsby, Andrei Giurgiu, Stanislaw Jastrzebski, Bruna Morrone, Quentin De~Laroussilhe, Andrea Gesmundo, Mona Attariyan, and Sylvain Gelly.
\newblock Parameter-efficient transfer learning for nlp.
\newblock In \emph{International conference on machine learning}, pp.\  2790--2799. PMLR, 2019.

\bibitem[Hu et~al.(2022)Hu, Shen, Wallis, Allen-Zhu, Li, Wang, Wang, Chen, et~al.]{hu2022lora}
Edward~J Hu, Yelong Shen, Phillip Wallis, Zeyuan Allen-Zhu, Yuanzhi Li, Shean Wang, Lu~Wang, Weizhu Chen, et~al.
\newblock Lora: Low-rank adaptation of large language models.
\newblock \emph{ICLR}, 1\penalty0 (2):\penalty0 3, 2022.

\bibitem[Hu et~al.(2023)Hu, Xie, Wang, Chen, and Pan]{hu2023structure}
Yahao Hu, Yifei Xie, Tianfeng Wang, Man Chen, and Zhisong Pan.
\newblock Structure-aware low-rank adaptation for parameter-efficient fine-tuning.
\newblock \emph{Mathematics}, 11\penalty0 (20):\penalty0 4317, 2023.

\bibitem[Jiang et~al.(2024)Jiang, Huang, Luo, Zhang, Huang, Wei, Deng, Sun, Zhang, Wang, et~al.]{jiang2024mora}
Ting Jiang, Shaohan Huang, Shengyue Luo, Zihan Zhang, Haizhen Huang, Furu Wei, Weiwei Deng, Feng Sun, Qi~Zhang, Deqing Wang, et~al.
\newblock Mora: High-rank updating for parameter-efficient fine-tuning.
\newblock \emph{arXiv preprint arXiv:2405.12130}, 2024.

\bibitem[Kaplan et~al.(2020)Kaplan, McCandlish, Henighan, Brown, Chess, Child, Gray, Radford, Wu, and Amodei]{kaplan2020scaling}
Jared Kaplan, Sam McCandlish, Tom Henighan, Tom~B Brown, Benjamin Chess, Rewon Child, Scott Gray, Alec Radford, Jeffrey Wu, and Dario Amodei.
\newblock Scaling laws for neural language models.
\newblock \emph{arXiv preprint arXiv:2001.08361}, 2020.

\bibitem[Lewis et~al.(2019)Lewis, Liu, Goyal, Ghazvininejad, Mohamed, Levy, Stoyanov, and Zettlemoyer]{lewis2019bart}
Mike Lewis, Yinhan Liu, Naman Goyal, Marjan Ghazvininejad, Abdelrahman Mohamed, Omer Levy, Ves Stoyanov, and Luke Zettlemoyer.
\newblock Bart: Denoising sequence-to-sequence pre-training for natural language generation, translation, and comprehension.
\newblock \emph{arXiv preprint arXiv:1910.13461}, 2019.

\bibitem[Liang et~al.(2021)Liang, Zuo, Chen, Jiang, Liu, He, Zhao, and Chen]{liang2021super}
Chen Liang, Simiao Zuo, Minshuo Chen, Haoming Jiang, Xiaodong Liu, Pengcheng He, Tuo Zhao, and Weizhu Chen.
\newblock Super tickets in pre-trained language models: From model compression to improving generalization.
\newblock \emph{arXiv preprint arXiv:2105.12002}, 2021.

\bibitem[Liu et~al.(2019)Liu, Ott, Goyal, Du, Joshi, Chen, Levy, Lewis, Zettlemoyer, and Stoyanov]{liu2019roberta}
Yinhan Liu, Myle Ott, Naman Goyal, Jingfei Du, Mandar Joshi, Danqi Chen, Omer Levy, Mike Lewis, Luke Zettlemoyer, and Veselin Stoyanov.
\newblock Roberta: A robustly optimized bert pretraining approach.
\newblock \emph{arXiv preprint arXiv:1907.11692}, 2019.

\bibitem[Mao et~al.(2024)Mao, Huang, Guan, Bao, Mo, and Xu]{mao2024dora}
Yulong Mao, Kaiyu Huang, Changhao Guan, Ganglin Bao, Fengran Mo, and Jinan Xu.
\newblock Dora: Enhancing parameter-efficient fine-tuning with dynamic rank distribution.
\newblock \emph{arXiv preprint arXiv:2405.17357}, 2024.

\bibitem[Mao et~al.(2025)Mao, Ge, Fan, Xu, Mi, Hu, and Gao]{mao2025survey}
Yuren Mao, Yuhang Ge, Yijiang Fan, Wenyi Xu, Yu~Mi, Zhonghao Hu, and Yunjun Gao.
\newblock A survey on lora of large language models.
\newblock \emph{Frontiers of Computer Science}, 19\penalty0 (7):\penalty0 197605, 2025.

\bibitem[Molchanov et~al.(2019)Molchanov, Mallya, Tyree, Frosio, and Kautz]{molchanov2019importance}
Pavlo Molchanov, Arun Mallya, Stephen Tyree, Iuri Frosio, and Jan Kautz.
\newblock Importance estimation for neural network pruning.
\newblock In \emph{Proceedings of the IEEE/CVF conference on computer vision and pattern recognition}, pp.\  11264--11272, 2019.

\bibitem[Narayan et~al.(2018)Narayan, Cohen, and Lapata]{narayan2018don}
Shashi Narayan, Shay~B Cohen, and Mirella Lapata.
\newblock Don't give me the details, just the summary! topic-aware convolutional neural networks for extreme summarization.
\newblock \emph{arXiv preprint arXiv:1808.08745}, 2018.

\bibitem[Paszke et~al.(2019)Paszke, Gross, Massa, Lerer, Bradbury, Chanan, Killeen, Lin, Gimelshein, Antiga, et~al.]{paszke2019pytorch}
Adam Paszke, Sam Gross, Francisco Massa, Adam Lerer, James Bradbury, Gregory Chanan, Trevor Killeen, Zeming Lin, Natalia Gimelshein, Luca Antiga, et~al.
\newblock Pytorch: An imperative style, high-performance deep learning library.
\newblock \emph{Advances in neural information processing systems}, 32, 2019.

\bibitem[Pfeiffer et~al.(2020)Pfeiffer, Kamath, R{\"u}ckl{\'e}, Cho, and Gurevych]{pfeiffer2020adapterfusion}
Jonas Pfeiffer, Aishwarya Kamath, Andreas R{\"u}ckl{\'e}, Kyunghyun Cho, and Iryna Gurevych.
\newblock Adapterfusion: Non-destructive task composition for transfer learning.
\newblock \emph{arXiv preprint arXiv:2005.00247}, 2020.

\bibitem[Radford et~al.(2019)Radford, Wu, Child, Luan, Amodei, Sutskever, et~al.]{radford2019language}
Alec Radford, Jeffrey Wu, Rewon Child, David Luan, Dario Amodei, Ilya Sutskever, et~al.
\newblock Language models are unsupervised multitask learners.
\newblock \emph{OpenAI blog}, 1\penalty0 (8):\penalty0 9, 2019.

\bibitem[Rajabzadeh et~al.(2024)Rajabzadeh, Valipour, Zhu, Tahaei, Kwon, Ghodsi, Chen, and Rezagholizadeh]{rajabzadeh2024qdylora}
Hossein Rajabzadeh, Mojtaba Valipour, Tianshu Zhu, Marzieh Tahaei, Hyock~Ju Kwon, Ali Ghodsi, Boxing Chen, and Mehdi Rezagholizadeh.
\newblock Qdylora: Quantized dynamic low-rank adaptation for efficient large language model tuning.
\newblock \emph{arXiv preprint arXiv:2402.10462}, 2024.

\bibitem[Rebuffi et~al.(2017)Rebuffi, Bilen, and Vedaldi]{rebuffi2017learning}
Sylvestre-Alvise Rebuffi, Hakan Bilen, and Andrea Vedaldi.
\newblock Learning multiple visual domains with residual adapters.
\newblock \emph{Advances in neural information processing systems}, 30, 2017.

\bibitem[Sanh et~al.(2020)Sanh, Wolf, and Rush]{sanh2020movement}
Victor Sanh, Thomas Wolf, and Alexander Rush.
\newblock Movement pruning: Adaptive sparsity by fine-tuning.
\newblock \emph{Advances in neural information processing systems}, 33:\penalty0 20378--20389, 2020.

\bibitem[Sui et~al.(2024)Sui, Yin, Gong, Xiao, Phan, and Yuan]{sui2024elrt}
Yang Sui, Miao Yin, Yu~Gong, Jinqi Xiao, Huy Phan, and Bo~Yuan.
\newblock Elrt: Efficient low-rank training for compact convolutional neural networks.
\newblock \emph{arXiv preprint arXiv:2401.10341}, 2024.

\bibitem[Valipour et~al.(2022)Valipour, Rezagholizadeh, Kobyzev, and Ghodsi]{valipour2022dylora}
Mojtaba Valipour, Mehdi Rezagholizadeh, Ivan Kobyzev, and Ali Ghodsi.
\newblock Dylora: Parameter efficient tuning of pre-trained models using dynamic search-free low-rank adaptation.
\newblock \emph{arXiv preprint arXiv:2210.07558}, 2022.

\bibitem[Vaswani et~al.(2017)Vaswani, Shazeer, Parmar, Uszkoreit, Jones, Gomez, Kaiser, and Polosukhin]{Vaswani+2017}
Ashish Vaswani, Noam Shazeer, Niki Parmar, Jakob Uszkoreit, Llion Jones, Aidan~N Gomez, \L~ukasz Kaiser, and Illia Polosukhin.
\newblock Attention is all you need.
\newblock In \emph{Advances in Neural Information Processing Systems}, volume~30. Curran Associates, Inc., 2017.
\newblock URL \url{https://proceedings.neurips.cc/paper_files/paper/2017/file/3f5ee243547dee91fbd053c1c4a845aa-Paper.pdf}.

\bibitem[Wang et~al.(2018)Wang, Singh, Michael, Hill, Levy, and Bowman]{wang2018glue}
Alex Wang, Amanpreet Singh, Julian Michael, Felix Hill, Omer Levy, and Samuel~R Bowman.
\newblock Glue: A multi-task benchmark and analysis platform for natural language understanding.
\newblock \emph{arXiv preprint arXiv:1804.07461}, 2018.

\bibitem[Wolf et~al.(2019)Wolf, Debut, Sanh, Chaumond, Delangue, Moi, Cistac, Rault, Louf, Funtowicz, et~al.]{wolf2019huggingface}
Thomas Wolf, Lysandre Debut, Victor Sanh, Julien Chaumond, Clement Delangue, Anthony Moi, Pierric Cistac, Tim Rault, R{\'e}mi Louf, Morgan Funtowicz, et~al.
\newblock Huggingface's transformers: State-of-the-art natural language processing.
\newblock \emph{arXiv preprint arXiv:1910.03771}, 2019.

\bibitem[Zaken et~al.(2021)Zaken, Ravfogel, and Goldberg]{zaken2021bitfit}
Elad~Ben Zaken, Shauli Ravfogel, and Yoav Goldberg.
\newblock Bitfit: Simple parameter-efficient fine-tuning for transformer-based masked language-models.
\newblock \emph{arXiv preprint arXiv:2106.10199}, 2021.

\bibitem[Zhai et~al.(2019)Zhai, Puigcerver, Kolesnikov, Ruyssen, Riquelme, Lucic, Djolonga, Pinto, Neumann, Dosovitskiy, et~al.]{zhai2019large}
Xiaohua Zhai, Joan Puigcerver, Alexander Kolesnikov, Pierre Ruyssen, Carlos Riquelme, Mario Lucic, Josip Djolonga, Andre~Susano Pinto, Maxim Neumann, Alexey Dosovitskiy, et~al.
\newblock A large-scale study of representation learning with the visual task adaptation benchmark.
\newblock \emph{arXiv preprint arXiv:1910.04867}, 2019.

\bibitem[Zhang et~al.(2023{\natexlab{a}})Zhang, Li, Chen, Jiang, Wang, and Qian]{zhang2023increlora}
Feiyu Zhang, Liangzhi Li, Junhao Chen, Zhouqiang Jiang, Bowen Wang, and Yiming Qian.
\newblock Increlora: Incremental parameter allocation method for parameter-efficient fine-tuning.
\newblock \emph{arXiv preprint arXiv:2308.12043}, 2023{\natexlab{a}}.

\bibitem[Zhang et~al.(2022)Zhang, Zuo, Liang, Bukharin, He, Chen, and Zhao]{zhang2022platon}
Qingru Zhang, Simiao Zuo, Chen Liang, Alexander Bukharin, Pengcheng He, Weizhu Chen, and Tuo Zhao.
\newblock Platon: Pruning large transformer models with upper confidence bound of weight importance.
\newblock In \emph{International conference on machine learning}, pp.\  26809--26823. PMLR, 2022.

\bibitem[Zhang et~al.(2023{\natexlab{b}})Zhang, Chen, Bukharin, Karampatziakis, He, Cheng, Chen, and Zhao]{zhang2023adalora}
Qingru Zhang, Minshuo Chen, Alexander Bukharin, Nikos Karampatziakis, Pengcheng He, Yu~Cheng, Weizhu Chen, and Tuo Zhao.
\newblock Adalora: Adaptive budget allocation for parameter-efficient fine-tuning.
\newblock \emph{arXiv preprint arXiv:2303.10512}, 2023{\natexlab{b}}.

\bibitem[Zhang et~al.(2024)Zhang, Qiang, Somayajula, and Xie]{zhang2024autolora}
Ruiyi Zhang, Rushi Qiang, Sai~Ashish Somayajula, and Pengtao Xie.
\newblock Autolora: Automatically tuning matrix ranks in low-rank adaptation based on meta learning.
\newblock \emph{arXiv preprint arXiv:2403.09113}, 2024.

\bibitem[Zhao et~al.(2024)Zhao, Zhang, Chen, Wang, Anandkumar, and Tian]{zhao2024galore}
Jiawei Zhao, Zhenyu Zhang, Beidi Chen, Zhangyang Wang, Anima Anandkumar, and Yuandong Tian.
\newblock Galore: Memory-efficient llm training by gradient low-rank projection.
\newblock \emph{arXiv preprint arXiv:2403.03507}, 2024.

\end{thebibliography}
\bibliographystyle{colm2025_conference}

\newpage
\appendix
\section{GLUE Dataset}
\label{app:glue}

The General Language Understanding Evaluation (GLUE) benchmark \citep{wang2018glue} is a widely used collection of natural language understanding (NLU) tasks designed to evaluate the performance of language models across diverse linguistic challenges. Below, we provide a brief description of each dataset included in GLUE.

\begin{table}[h]
    \centering
    \caption{Summary of the GLUE Benchmark}
    \label{tab:glue_summary}
    \begin{tabular}{l c c c l}
        \toprule
        \textbf{Corpus} & \textbf{\#Train} & \textbf{\#Dev} & \textbf{\#Test} & \textbf{Metric} \\
        \midrule
        \multicolumn{5}{c}{\textit{Single-Sentence Classification}} \\
        CoLA & 8.5k & 1k & 1k & Matthews corr \\
        SST-2 & 67k & 872 & 1.8k & Accuracy \\
        \midrule
        \multicolumn{5}{c}{\textit{Pairwise Text Classification}} \\
        MNLI & 393k & 20k & 20k & Accuracy \\
        RTE & 2.5k & 276 & 3k & Accuracy \\
        QQP & 364k & 40k & 391k & Accuracy/F1 \\
        MRPC & 3.7k & 408 & 1.7k & Accuracy/F1 \\
        QNLI & 108k & 5.7k & 5.7k & Accuracy \\
        \midrule
        \multicolumn{5}{c}{\textit{Text Similarity}} \\
        STS-B & 7k & 1.5k & 1.4k & Pearson/Spearman corr \\
        \bottomrule
    \end{tabular}
\end{table}

\begin{itemize}
    \item \textbf{MNLI (Multi-Genre Natural Language Inference)}  \\
    \textbf{Task:} Sentence-pair classification task where a model determines whether a hypothesis is \textit{entailed by}, \textit{contradictory to}, or \textit{neutral} with respect to a given premise.  \\
    \textbf{Metric:} Accuracy (evaluated separately on \textit{matched} and \textit{mismatched} sections of the dataset).  

    \item \textbf{SST-2 (Stanford Sentiment Treebank)}  \\
    \textbf{Task:} Binary classification task for determining whether a sentence expresses a \textit{positive} or \textit{negative} sentiment.  \\
    \textbf{Metric:} Accuracy.  

    \item \textbf{MRPC (Microsoft Research Paraphrase Corpus)}  \\
    \textbf{Task:} Binary classification task to determine whether two sentences are \textit{paraphrases} of each other.  \\
    \textbf{Metric:} Accuracy and F1 score.  

    \item \textbf{CoLA (Corpus of Linguistic Acceptability)}  \\
    \textbf{Task:} Binary classification task to predict whether a sentence is \textit{linguistically acceptable}.  \\
    \textbf{Metric:} Matthews correlation coefficient (MCC).  

    \item \textbf{QNLI (Question Natural Language Inference)}  \\
    \textbf{Task:} Binary classification task to determine whether a given context sentence \textit{contains the answer} to a question.  \\
    \textbf{Metric:} Accuracy.  

    \item \textbf{QQP (Quora Question Pairs)}  \\
    \textbf{Task:} Binary classification task to determine whether two questions on Quora are \textit{semantically equivalent}.  \\
    \textbf{Metric:} Accuracy and F1 score.  

    \item \textbf{RTE (Recognizing Textual Entailment)}  \\
    \textbf{Task:} Binary classification task to determine whether a hypothesis is \textit{entailed} by a given premise.  \\
    \textbf{Metric:} Accuracy.  

    \item \textbf{STS-B (Semantic Textual Similarity Benchmark)}  \\
    \textbf{Task:} Regression task to predict a similarity score between two sentences, ranging from 0 (completely dissimilar) to 5 (semantically equivalent).  \\
    \textbf{Metric:} Pearson and Spearman correlation coefficients.  
\end{itemize}

\section{Natural Language Understanding Training Configurations}
\label{app:nlu_config}
\begin{table*}[h!]
    \centering
    \tiny
    \begin{tabular}{lcccccccc}
        \hline
        Corpus & batch size & learning rate & epochs & $t_{warmup}$ & $t_{stabilize}$ & $t_{adjust\_interval}$ & b (r=2/4/10) & k (r=2/4/10) \\
        \hline
        MNLI & 64 & 5e-04 & 5 & 5000 & 10000 & 500 & 1/2/4 & 1/1/2 \\
        SST-2 & 64 & 8e-04 & 20 & 3000 & 10000 & 500 & 1/2/4 & 1/1/2 \\
        CoLA & 64 & 8e-04 & 20 & 300 & 1500 & 50 & 1/2/4 & 1/1/2 \\
        QQP & 128 & 4e-04 & 5 & 3000 & 5000 & 100 & 1/2/4 & 1/1/2 \\
        QNLI & 64 & 5e-04 & 5 & 2000 & 2000 & 200 & 1/2/4 & 1/1/2 \\
        RTE & 64 & 1.2e-03 & 50 & 300 & 500 & 50 & 1/4/8 & 1/2/4 \\
        MRPC & 64 & 5e-04 & 30 & 300 & 400 & 50 & 2/2/4 & 1/1/2 \\
        STS-B & 64 & 5e-03 & 25 & 300 & 300 & 150 & 3/4/8 & 1/2/4 \\
        \hline
    \end{tabular}
    \vspace{10pt}
    \caption{Hyperparameters used for different GLUE tasks}
    \label{tab:hyperparams}
\end{table*}

\section{Natural Language Generating Training Configurations}
\label{app:nlg_config}
\begin{table*}[h!]
    \centering
    \tiny
    \begin{tabular}{lcccccccc}
        \hline
        Corpus & batch size & learning rate & epochs & $t_{warmup}$ & $t_{stabilize}$ & $t_{adjust\_interval}$ & b (r=2/6) & k (r=2/6) \\
        \hline
        XSum & 32 & 5e-04 & 2 & 1500 & 3000 & 200 & 2/3 & 1/2 \\
        \hline
    \end{tabular}
    \vspace{10pt}
    \caption{Hyperparameters used for XSum task.}
\end{table*}

\section{VTAB Dataset}
\label{app:vtab_dataset}

The Visual Task Adaptation Benchmark (VTAB) \cite{zhai2019large} is a diverse suite of 19 vision tasks drawn from public datasets, designed to assess the transferability and generalization of visual representations. VTAB is organized into three categories:

\begin{itemize}
    \item \textbf{Natural:} Tasks based on real-world images that capture everyday objects and scenes. Examples include Caltech101, CIFAR-100, DTD, 102 Category Flower, Oxford-IIIT Pet, SUN397, and SVHN.
    \item \textbf{Specialized:} Tasks focusing on domain-specific images that often have limited samples and unique characteristics. This category includes Diabetic Retinopathy, EuroSAT, KITTI distance prediction, and PatchCamelyon.
    \item \textbf{Structured:} Tasks that involve synthetic or controlled environments, where understanding spatial relationships is key. Notable datasets here are CLEVR distance prediction, CLEVR counting, DMLab Frames, dSprites orientation and location prediction, and Small NORB (azimuth and elevation prediction).
\end{itemize}
See Table \ref{tab:vtab_summary} for different datasets' description. For each category, we randomly selected several datasets to ensure a balanced evaluation of model performance across different visual challenges.

\begin{table}[h]
    \centering
    \caption{Summary of the VTAB Benchmark}
    \label{tab:vtab_summary}
    \begin{tabular}{l c l}
        \toprule
        \textbf{Dataset} & \textbf{Category} & \textbf{Task Description} \\
        \midrule
        Caltech101 & Natural & Object recognition \\
        CIFAR-100 & Natural & Image classification \\
        CLEVR distance prediction & Structured & Distance estimation in synthetic scenes \\
        CLEVR counting & Structured & Object counting in synthetic scenes \\
        Diabetic Retinopathy & Specialized & Disease severity assessment \\
        DMLab Frames & Structured & Analysis of environment frames \\
        dSprites orientation prediction & Structured & Predict object orientation \\
        dSprites location prediction & Structured & Predict object location \\
        DTD & Natural & Texture classification \\
        EuroSAT & Specialized & Land use classification from satellite imagery \\
        KITTI distance prediction & Specialized & Distance estimation in driving scenes \\
        102 Category Flower & Natural & Flower species classification \\
        Oxford-IIIT Pet & Natural & Pet breed classification \\
        PatchCamelyon & Specialized & Detection of cancerous regions in histopathology \\
        Resisc45 & Specialized & Land use classification \\
        Small NORB azimuth prediction & Structured & Azimuth angle prediction \\
        Small NORB elevation prediction & Structured & Elevation angle prediction \\
        SUN397 & Natural & Scene recognition \\
        SVHN & Natural & Digit recognition in street view images \\
        \bottomrule
    \end{tabular}
\end{table}

\section{Visual Task Training Configurations}
\label{app:vit_config}
\begin{table}[H]
    \centering
    \tiny
    \begin{tabular}{lcccccccc}
        \hline
        Corpus & batch size & learning rate & epochs & $t_{warmup}$ & $t_{stabilize}$ & $t_{adjust\_interval}$ & b/k  \\
        \hline
        CIFAR & 16 & 1e-03 & 100 & 500 & 1500 & 200 & 6/3 \\
        SVHN & 16 & 1e-03 & 100 & 500 & 1500 & 200 & 4/2 \\
        Eurosat & 16 & 1e-03 & 100 & 500 & 1500 & 200 & 4/2 \\
        Resisc45 & 16 & 1e-03 & 100 & 500 & 1500 & 200 & 6/3 \\
        dSpr\-Ori & 16 & 1e-03 & 100 & 500 & 1500 & 200 & 4/2 \\
        DMLab & 16 & 1e-03 & 100 & 500 & 1500 & 200 & 6/3 \\
        \hline
    \end{tabular}
    \vspace{10pt}
    \caption{Hyperparameters used for different VTAB tasks}
    \label{tab:vit_hyperparams}
\end{table}

\section{Final Rank Heatmap}
\label{app:heatmap}
\begin{figure}[h!]
    \centering
    \begin{minipage}{0.49\textwidth}
        \centering
        \includegraphics[width=\linewidth]{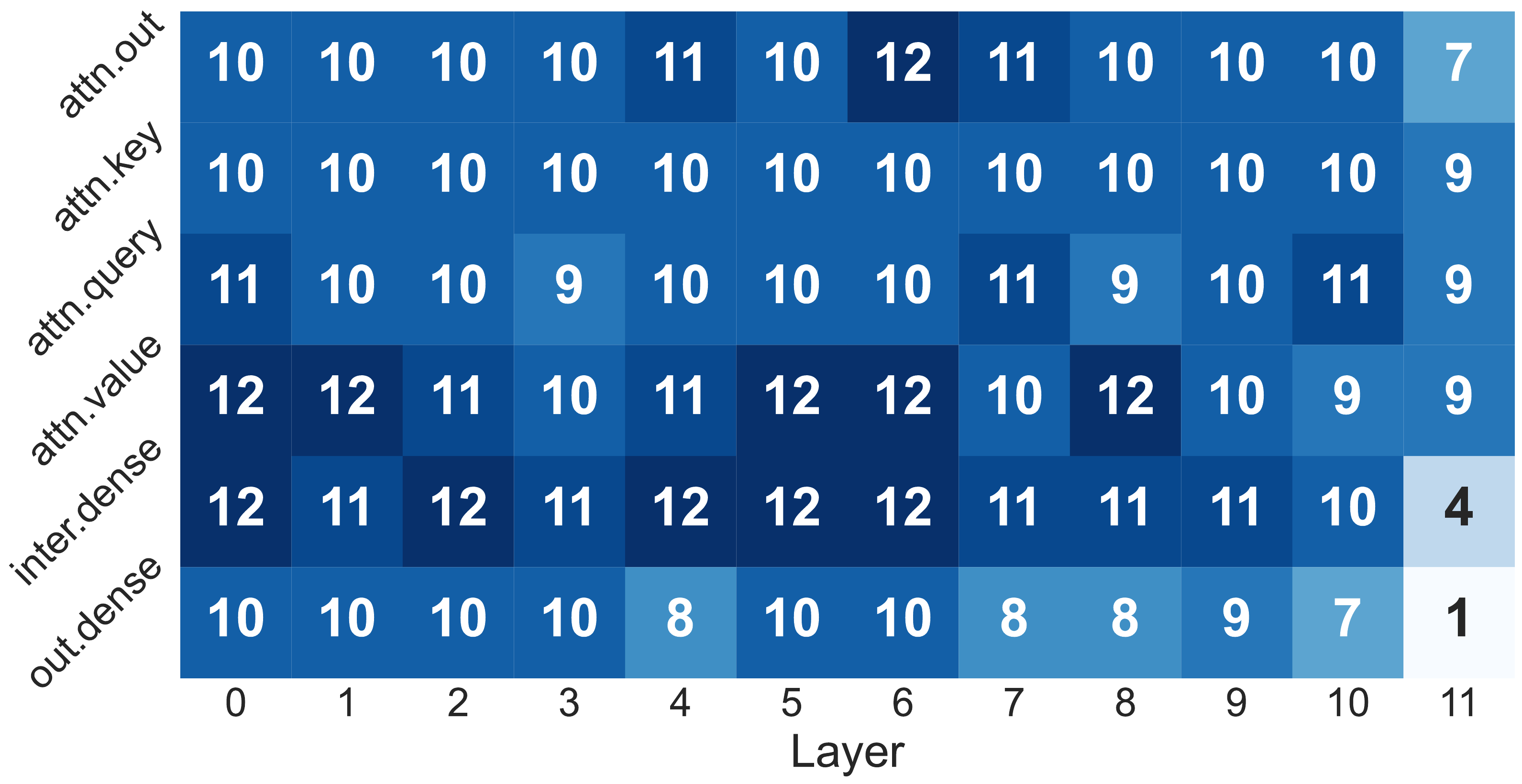}
    \end{minipage}
    \hfill
    \begin{minipage}{0.49\textwidth}
        \centering
        \includegraphics[width=\linewidth]{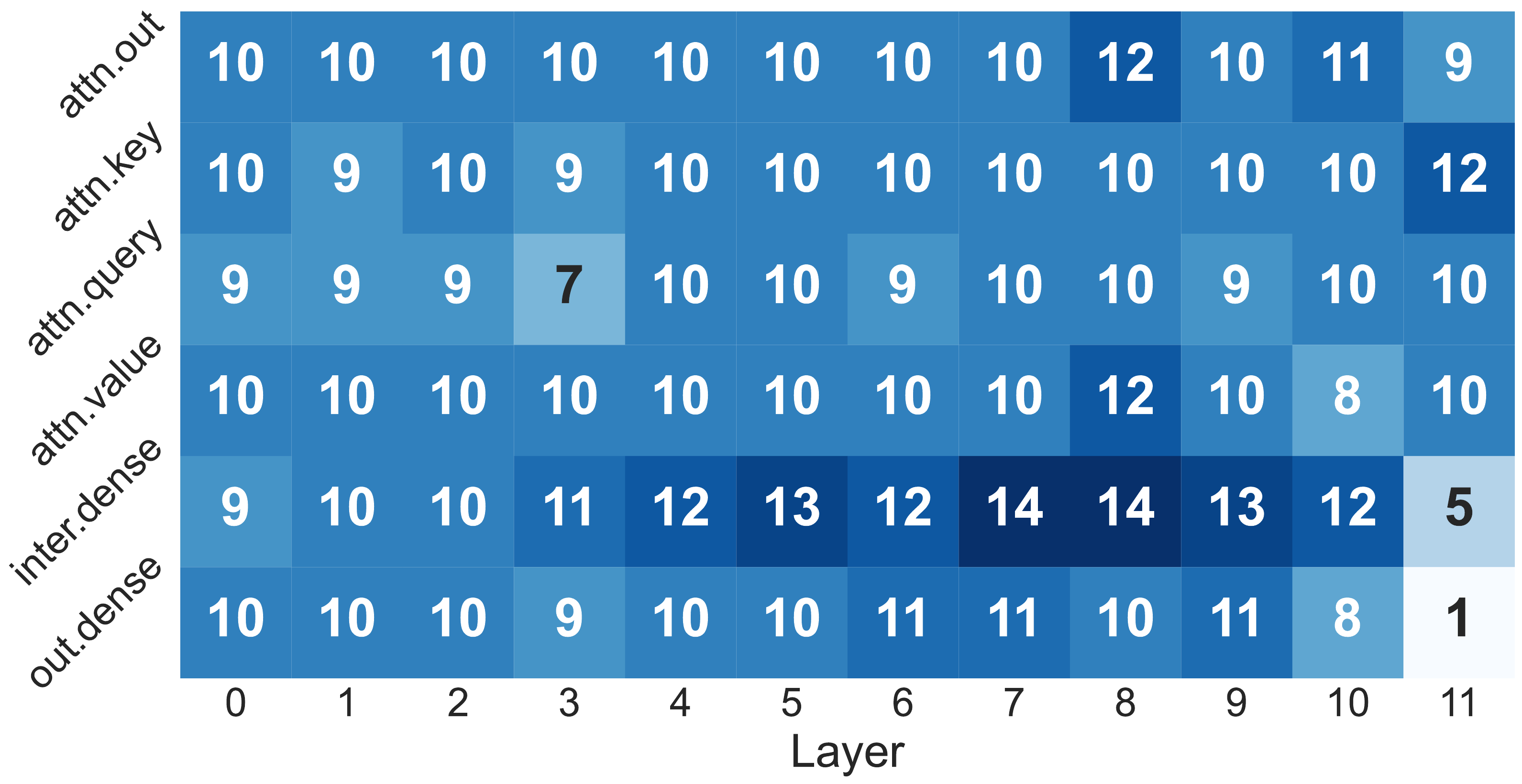}
    \end{minipage}
    \caption{MNLI (Left) vs MRPC (right) Final Rank Heatmap with $r=10$}
    \label{fig:mnli_mrpc}
\end{figure}

Figure \ref{fig:mnli_mrpc} gives the final rank heatmap by ElaLoRA with rank 10. We can tell that MNLI has higher ranks in the front layers while MRPC shows higher ranks in the later layers. Intuitively, in the MNLI task, which involves deep syntactic and semantic analysis of sentence relationships, higher ranks in the front layers are crucial for capturing fine-grained features and initial representations early on. Conversely, in the MRPC task, which focuses on paraphrase detection through higher-level semantic comparison, higher ranks in the later layers are more important for integrating and comparing semantic representations effectively. This reflects the success of ElaLoRA in addressing the differing demands of the tasks: MNLI requires detailed initial processing, while MRPC relies on robust later-stage integration and comparison.

\begin{figure}[h]
    \centering
    \begin{minipage}{0.32\textwidth}
        \centering
        \includegraphics[width=\linewidth]{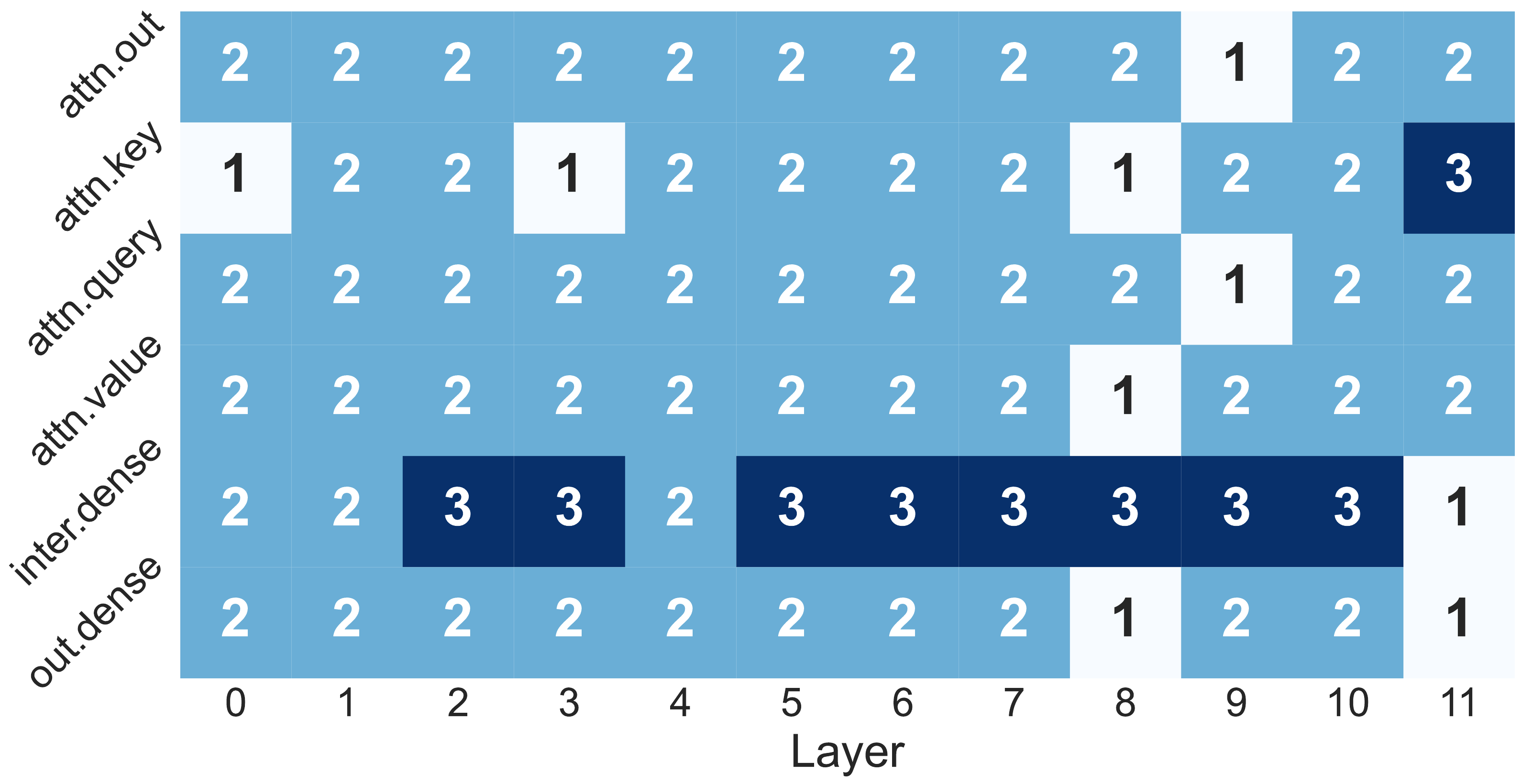}
    \end{minipage}
    \hfill
    \begin{minipage}{0.32\textwidth}
        \centering
        \includegraphics[width=\linewidth]{Images/RTE_ela_r4.pdf}
    \end{minipage}
    \hfill
    \begin{minipage}{0.32\textwidth}
        \centering
        \includegraphics[width=\linewidth]{Images/RTE_ela_r10.pdf}
    \end{minipage}
    \caption{RTE Final Rank Heatmap with $r=2$ (left)$/4$ (middle)$/10$ (right)}
    \label{fig:rte}
\end{figure}

\begin{figure}[h]
    \centering
    \begin{minipage}{0.32\textwidth}
        \centering
        \includegraphics[width=\linewidth]{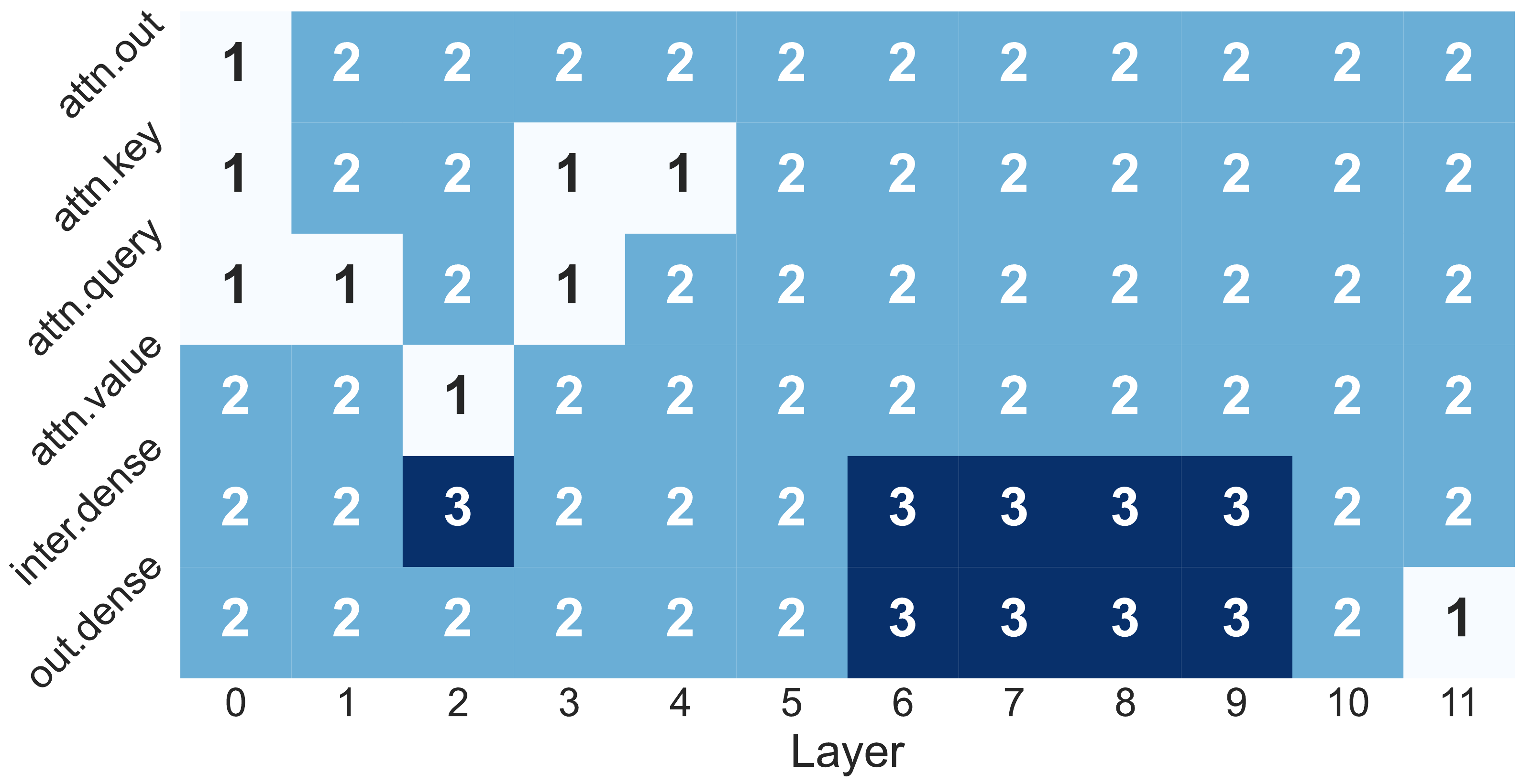}
    \end{minipage}
    \hfill
    \begin{minipage}{0.32\textwidth}
        \centering
        \includegraphics[width=\linewidth]{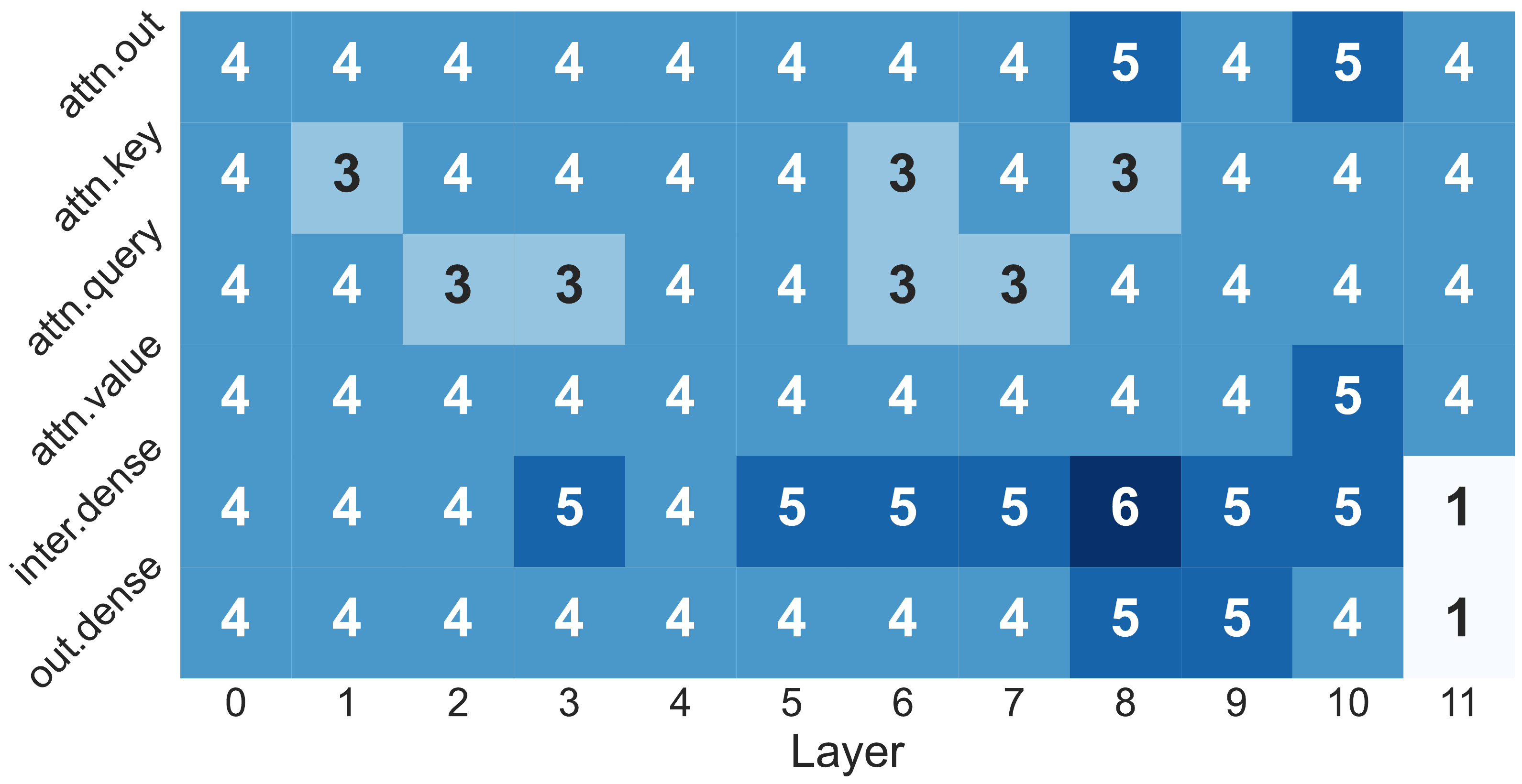}
    \end{minipage}
    \hfill
    \begin{minipage}{0.32\textwidth}
        \centering
        \includegraphics[width=\linewidth]{Images/Appendix_Rank_Plots/ela_r10_MRPC.pdf}
    \end{minipage}
    \caption{MRPC Final Rank Heatmap with $r=2$ (left)$/4$ (middle)$/10$ (right)}
    \label{fig:mrpc}
\end{figure}

\begin{figure}[h]
    \centering
    \begin{minipage}{0.32\textwidth}
        \centering
        \includegraphics[width=\linewidth]{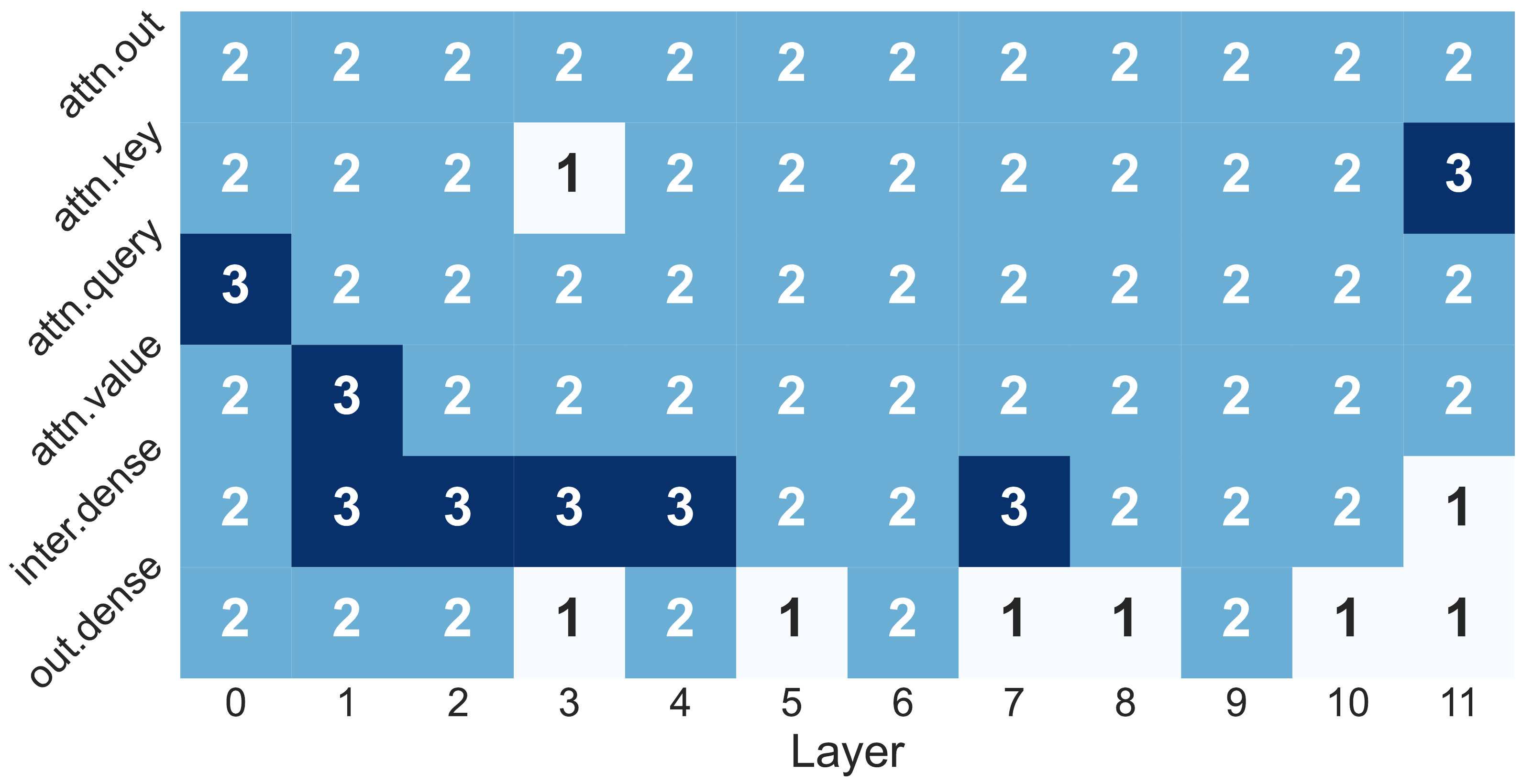}
    \end{minipage}
    \hfill
    \begin{minipage}{0.32\textwidth}
        \centering
        \includegraphics[width=\linewidth]{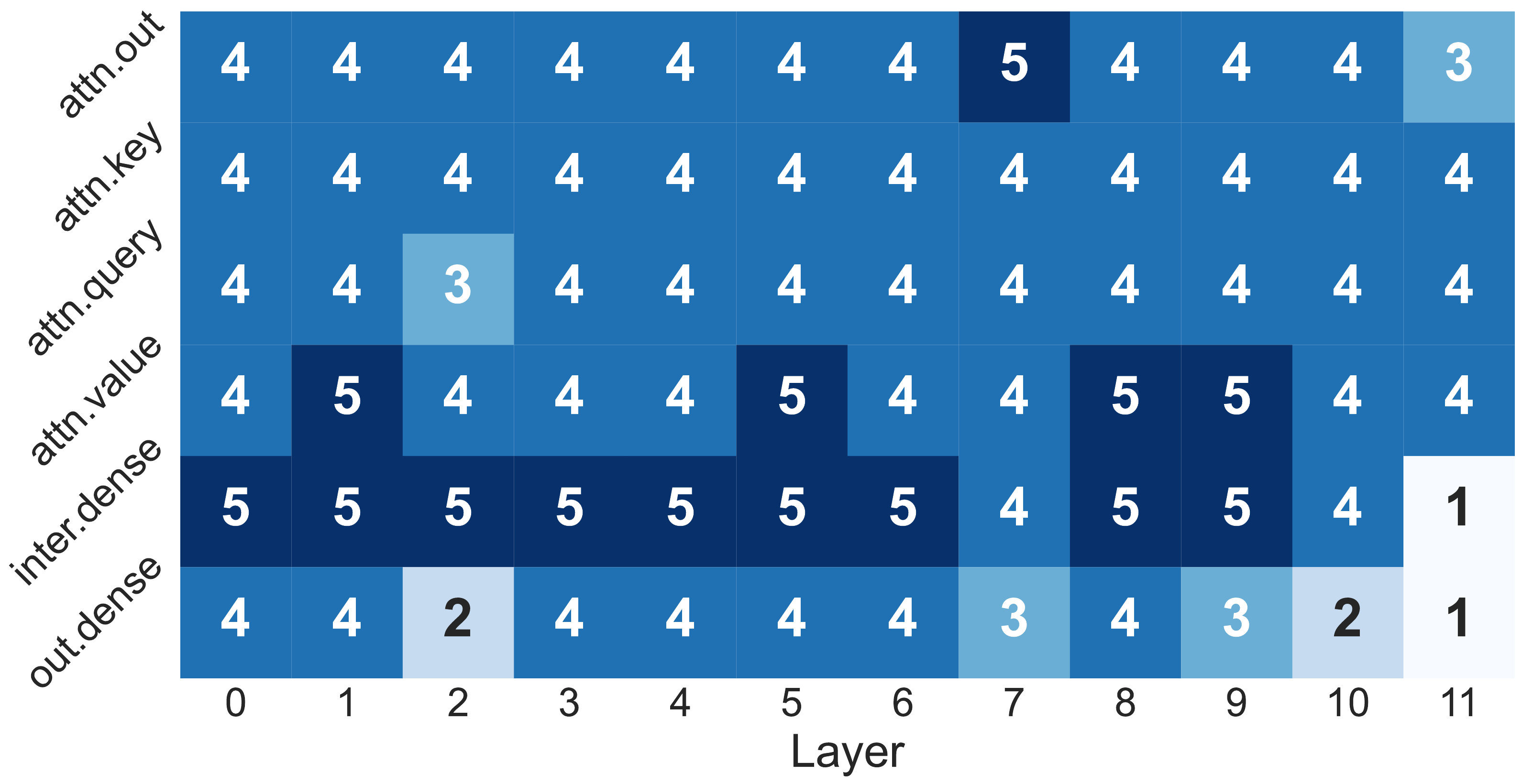}
    \end{minipage}
    \hfill
    \begin{minipage}{0.32\textwidth}
        \centering
        \includegraphics[width=\linewidth]{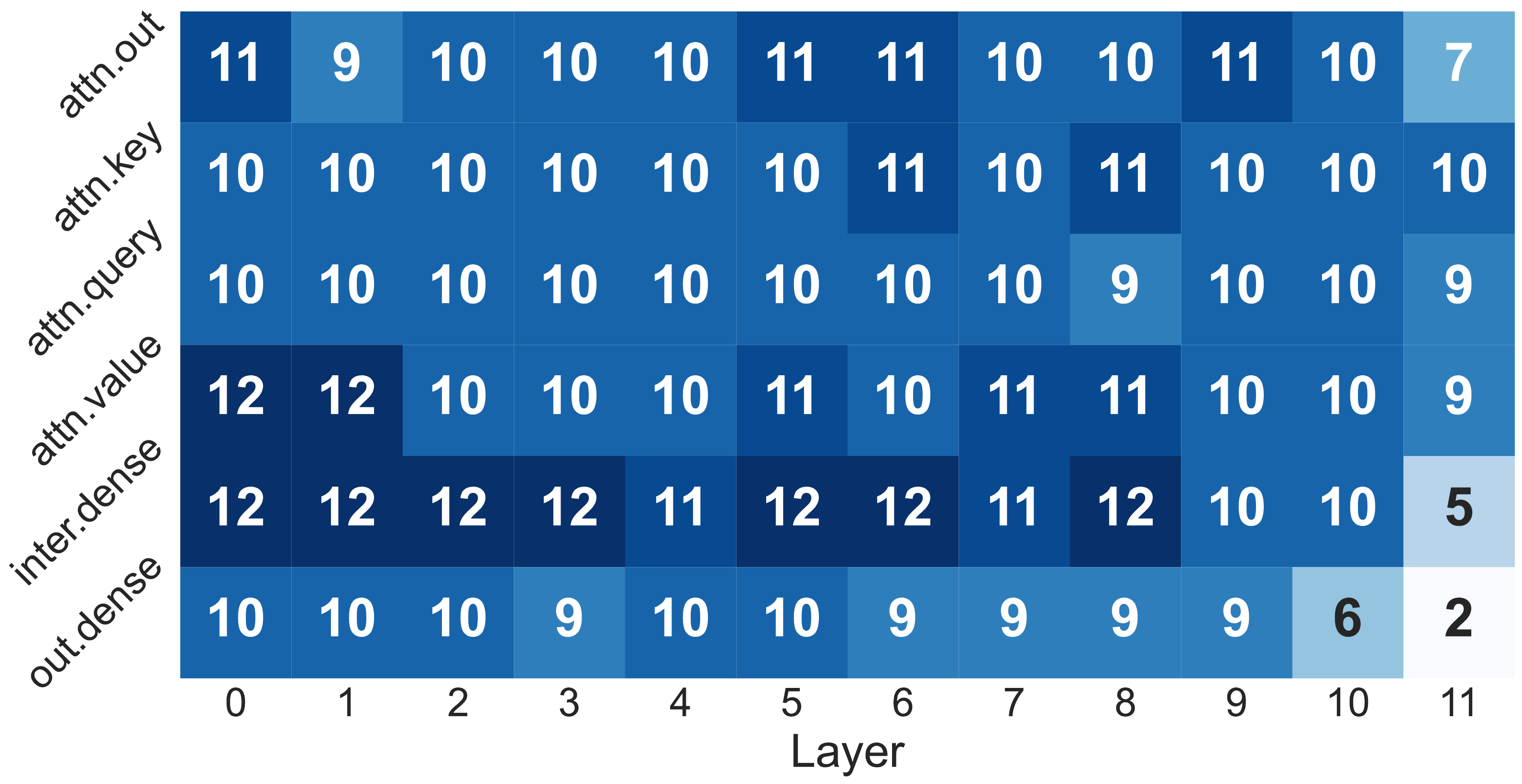}
    \end{minipage}
    \caption{SST2 Final Rank Heatmap with $r=2$ (left)$/4$ (middle)$/10$ (right)}
    \label{fig:sst2}
\end{figure}

Figures \ref{fig:rte}, \ref{fig:mrpc}, and \ref{fig:sst2} present the final rank heatmaps generated by ElaLoRA. These heatmaps demonstrate that, regardless of the rank settings, ElaLoRA consistently allocates ranks across various layers and weight matrices in a similar manner. This consistency highlights the robustness and reliability of ElaLoRA in adapting to different rank configurations.

\end{document}